\documentclass[10pt,twocolumn,letterpaper]{article}

\usepackage[pagenumbers]{iccv} 
\usepackage{times}
\usepackage{epsfig}
\usepackage{graphicx}
\usepackage{amsmath}
\usepackage{amssymb}
\usepackage{booktabs}
\usepackage{mathrsfs}
\usepackage{caption}
\def\name{SAS}
\definecolor{iccvblue}{rgb}{0.21,0.49,0.74}
\usepackage[pagebackref,breaklinks,colorlinks,allcolors=iccvblue]{hyperref}

\title{SAS: Segment Any 3D Scene with Integrated 2D Priors}

\author{Zhuoyuan Li$^1$\footnotemark[1]~, Jiahao Lu$^1$\footnotemark[1]~, Jiacheng Deng$^1$,Hanzhi Chang$^1$, \\ Lifan Wu$^1$, Yanzhe Liang$^1$, Tianzhu Zhang$^{1}$\footnotemark[2]~ \\
\small{$^1$University of Science and Technology of China}\\
\small{Project Page: \url{https://peoplelu.github.io/SAS.github.io}} \\
}

\begin{document}
\twocolumn[{%
\renewcommand\twocolumn[1][]{#1}%
\maketitle
\includegraphics[width=\linewidth]{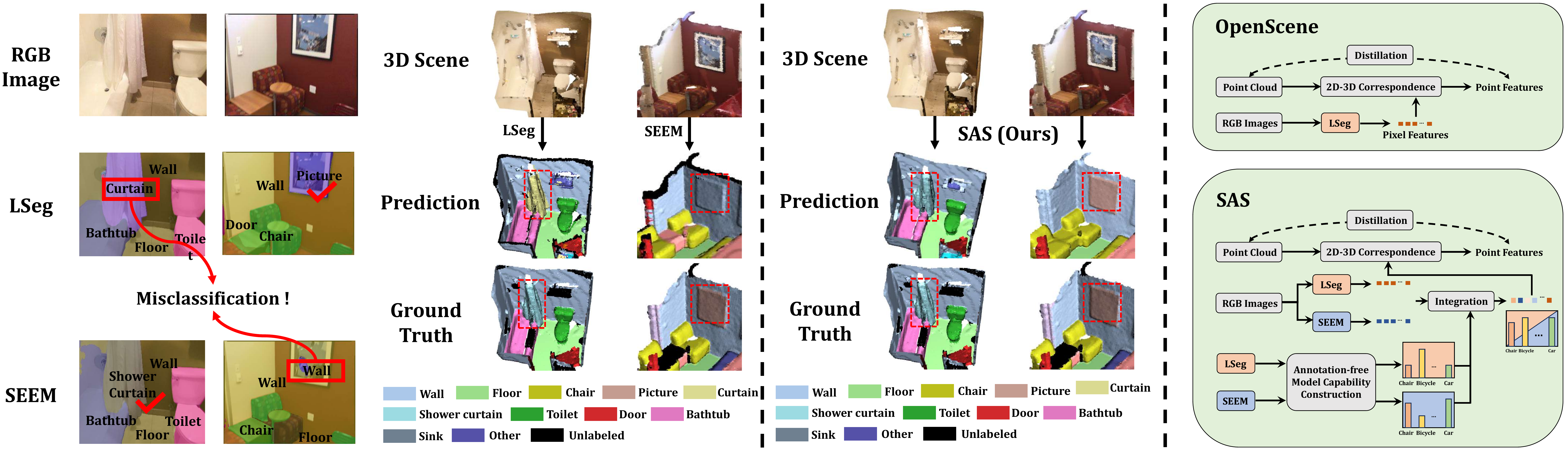}
\captionof{figure}{\textbf{Left}: The leading 2D open vocabulary models like LSeg \cite{lseg} and SEEM \cite{SEEM} often misidentify objects, which makes the distilled 3D model perform the same misidentification.
  \textbf{Middle}: Our proposed \name{} successfully correct the misidentified object.
  \textbf{Right}: \name{} distills open vocabulary knowledge from multiple 2D models with novel designs, e.g., Annotation-free Model Capability Construction.}
\label{fig:intro}
\vspace{19pt}
}]

\renewcommand{\thefootnote}{\fnsymbol{footnote}}
\footnotetext[1]{Equal Contribution}
\footnotetext[2]{Corresponding Author}
\begin{abstract}
    The open vocabulary capability of 3D models is increasingly valued, as traditional methods with models trained with fixed categories fail to recognize unseen objects in complex dynamic 3D scenes. 
In this paper, we propose a simple yet effective approach, \textbf{\name{}}, to integrate the open vocabulary capability of multiple 2D models and migrate it to 3D domain. Specifically, we first propose \textbf{Model Alignment via Text} to map different 2D models into the same embedding space using text as a bridge. Then, we propose \textbf{Annotation-Free Model Capability Construction} to explicitly quantify the 2D model's capability of recognizing different categories using diffusion models. Following this, point cloud features from different 2D models are fused with the guide of constructed model capabilities. Finally, the integrated 2D open vocabulary capability is transferred to 3D domain through feature distillation. \name{} outperforms previous methods by a large margin across multiple datasets, including ScanNet v2, Matterport3D, and nuScenes, while its generalizability is further validated on downstream tasks, e.g., gaussian segmentation and instance segmentation.
 
\end{abstract}

\section{Introduction}

3D scene understanding is a fundamental task that aims to predict the semantics for every 3D point. Numerous real-world applications, such as autonomous driving \cite{nuscenes, kitti, waymo}, virtual reality \cite{VR}, and robot manipulation \cite{robot}, depend heavily on 3D scene understanding. Traditional methods in this area perform supervised training on labeled 3D datasets with a limited number of categories, which hinders the model's ability to identify unseen objects. This prompts researchers to turn their attention to open vocabulary capabilities of 3D scene understanding models.

In 2D open vocabulary understanding, a typical approach is to align image features to language features by contrast learning on a large number of image-text pairs, e.g., CLIP \cite{clip}. By analogy, contrast learning on a large number of point-text pairs is supposed to enable 3D open vocabulary understanding. However, point-text pairs are much harder to acquire than image-text pairs due to the high cost and time-consuming nature of point cloud annotation \cite{pla, openscene}. 
A compromise approach is to distill the open vocabulary capability of 2D open vocabulary models onto the 3D models \cite{pla, regionplc, openscene, clip2scene, clipfo3d, GGSD, diff2scene, ov3d}. For example, OpenScene \cite{openscene} adopts 2D open vocabulary segmentaters to extract pixel features for point clouds to enable point-language alignment. These pixel features are distilled to transfer the 2D open vocabulary capability to 3D models.

Although distillation-based methods have become dominant in 3D scene understanding, the distilled model inherited its 3D open vocabulary capability from the adopted 2D models, making the performance of the distilled model vary with the performance of the adopted 2D model. As shown in Fig. \ref{fig:intro}, the 2D open vocabulary model may incorrectly recognize objects in the scene, which generates ambiguous supervision signals for the 3D model so that the 3D model will make the same mistake. For example, SEEM \cite{SEEM} misidentifies a picture as a wall (left), making the distilled model perform the same misidentification (left). Therefore, an intuitive idea is to integrate the capabilities of different 2D models to correct misidentified objects.
However, there are two inherent difficulties in integrating the capability of different 2D models. First, different 2D models have unaligned image-text feature spaces. The unaligned features cannot be directly integrated. Second, testing the model on a batch of images reflects the capability of this model. However, since 3D open vocabulary tasks are inherently designed to recognize unseen categories in a zero-shot fashion, obtaining test images and their annotations is challenging. As a result, it is impractical to directly evaluate the model’s performance.

To address the above problems, we propose \textbf{\name{}}, to learn better 3D representations from the integration of multiple 2D models. Our key idea is to explicitly model 2D models' capability of identifying different categories in the scene, and use this to guide the integration of different 2D models, which produces better 3D representations than a single 2D model can do.
First, we propose \textbf{Model Alignment via Text} to align features of different 2D models using text as a bridge. Specifically, a 2D open vocabulary model, e.g., SEEM model \cite{SEEM}, predicts a label for each pixel (or mask), based on which we further generate a caption containing more complex information such as color, shape, etc., for each pixel (or mask), which enriches the semantic information and intra-class diversity. Now, different 2D open vocabulary models are aligned at the text-level. The caption is then inputted into a shared text encoder to get the aligned features.
Second, we propose \textbf{Annotation-free Model Capability Construction} to quantitatively measure the capability of different 2D open vocabulary models. Specifically, it leverages the text-to-image diffusion model \cite{stablediffusion} to synthesize images of common types. We utilize the 2D open vocabulary model's performance on these synthesized images to quantify the model's capabilities, which later can guide the integration of different 2D models.
Finally, \name{} transfers the integrated open vocabulary capability of 2D models to 3D domain via feature distillation. 

\begin{figure*}[t]
  \centering
  \includegraphics[scale=0.49]{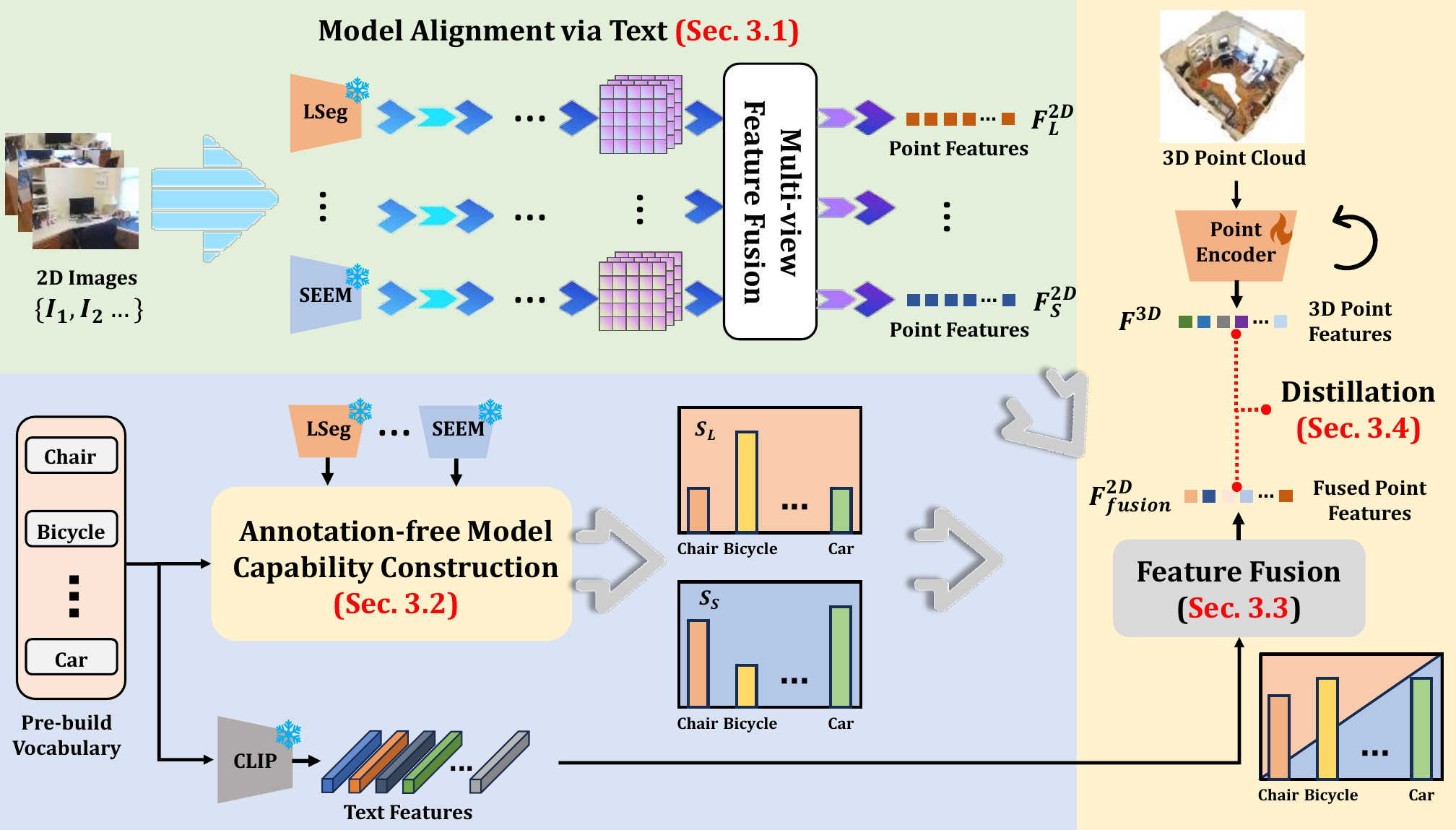}
  \caption{
  \textbf{Overview of our proposed \name{}.} \name{} first align features of different models in a unified embedding space (Sec. \ref{sec:3.1}). Then \name{} constructs models' capability to recognize various objects (Sec. \ref{sec:3.2}). With the constructed capability as guide, features from different 2D models are integrated (Sec. \ref{sec:3.3}). Finally, a 3D network is distilled to enable 3D open vocabulary understanding (Sec. \ref{sec:3.4}). 
  }
  \label{fig:overall}
  \vspace{-10pt}
\end{figure*}

We evaluate \name{} on multiple datasets, including ScanNet v2 \cite{scannet}, Matterport3D \cite{matterport3d} and nuScenes \cite{nuscenes}. Overall, we make the following contributions:
\begin{itemize}
  \item    
    We propose \name{}, the first approach to learn better 3D representations from multiple 2D models.
  \item 
    We propose Model Alignment via Text and Annotation-free Model-capability Construction to address the aforementioned difficulties in integrating different 2D models. The former aligns the features of models with different embedding spaces, while the latter explicitly models the model capability via synthetic images, which is then used to guide the integration of different model features.
  \item 
    We validate \name{} on different 3D scene understanding tasks, including semantic segmentation, instance segmentation, and gaussian segmentation. The results show that \name{} significantly outperforms previous methods while preserving strong generalizability.
\end{itemize}

\section{Related Work}
\textbf{Closed-set 3D Scene Understanding.} Recent work on closed-set 3D scene understanding has seen significant advancement, spanning tasks such as 3D semantic segmentation~\cite{qi2017pointnet++,qi2017pointnet,landrieu2018large,zhao2021point,wu2024point,li2024mamba24}, object detection~\cite{zhou2018voxelnet,lang2019pointpillars,pan20213d,shen2023v,deng2024diff3detr,kolodiazhnyi2024oneformer3d}, instance segmentation~\cite{jiang2020pointgroup,vu2022softgroup,sun2023superpoint,lu2023query,lai2023mask,kolodiazhnyi2024oneformer3d,lu2025beyond}, and shape correspondence~\cite{groueix20183d,deprelle2019learning,lang2021dpc,zeng2021corrnet3d,deng2023se,deng2024unsupervised}. Some studies~\cite{li2023mseg3d,wang2023multi,li2022deepfusion,tang2022bi,deng2025quantity} are exploring the use of multimodal information, particularly 2D images, to enhance closed-set 3D scene understanding. However, these methods primarily focus on supervised 3D tasks rather than open-vocabulary problems. Our approach aims to better leverage and integrate existing 2D open-vocabulary foundation models to achieve more accurate 3D open-vocabulary scene understanding tasks.

\textbf{Open-Vocabulary 2D Scene Understanding.}
The rapid development of vision-language models, such as CLIP~\cite{radford2021learning}, has driven zero-shot 2D scene understanding tasks. However, image-level recognition falls short for real-world applications. As a result, many works~\cite{li2022language,ghiasi2022scaling,zou2024segment,liang2023open,xu2023side} aim to correlate pixel embeddings with text embeddings to enable dense prediction tasks like segmentation. LSeg~\cite{li2022language} and OpenSeg~\cite{ghiasi2022scaling} are pioneering works in 2D open-vocabulary semantic segmentation via pixel-level text alignment. Following OpenScene~\cite{peng2023openscene}, we use 2D open-vocabulary segmentation models as teacher models for distilling 3D models. However, unlike previous works~\cite{peng2023openscene,wang2024open,zhu2024open,jiang2024open}, we find that different 2D open-vocabulary segmentation models exhibit varying recognition capabilities across categories. To better utilize these models, our method quantitatively evaluates their performance on annotation-free categories and integrates their strengths for improved scene understanding.

\textbf{Open-Vocabulary 3D Scene Understanding.}
Recent works on open-vocabulary 3D scene understanding can be classified into two categories based on representation patterns. The first category~\cite{kerr2023lerf,cen2023segment,qin2024langsplat,ye2024gaussian}, represented by works like LERF~\cite{kerr2023lerf} and LangSplat~\cite{qin2024langsplat}, distills 2D features into NeRF~\cite{mildenhall2021nerf} or 3DGS~\cite{kerbl20233d}. However, generating NeRF or 3DGS in a feed-forward manner remains challenging. The second category~\cite{peng2023openscene,wang2024open,zhu2024open,jiang2024open}, including methods like OpenScene~\cite{peng2023openscene}, bridges the gap between point clouds and images using camera parameters and depth. OV3D~\cite{jiang2024open} enriches textual descriptions with contextual information from foundation models, enabling richer feature distillation during alignment. Diff2Scene~\cite{zhu2024open} utilizes text-image generative models~\cite{rombach2022high} for open-vocabulary 3D scene understanding, while GGSD~\cite{wang2024open} leverages the Mean-teacher paradigm~\cite{tarvainen2017mean} for improved distillation. According to these methods, we note that the current capability of open-vocabulary 3D scene understanding heavily depends on the scene understanding ability of 2D foundation models~\cite{li2022language,ghiasi2022scaling,zou2024segment,liang2023open,xu2023side}. Therefore, exploring the potential of these 2D models is crucial for advancing 3D scene understanding. Based on this, our method tries to acquire a quantitative evaluation of their performance across different categories and integrate their strengths for better scene understanding.

\section{Method}

An overview of \name{} is illustrated in Fig. \ref{fig:overall}. Our initial step involves aligning features of different 2D models to the same feature space. The 2D features are then aggregated onto 3D points through pixel-point correspondence (Sec. \ref{sec:3.1}). We then utilize diffusion models to quantify 2D models' capabilities of identifying different categories in the scene (Sec. \ref{sec:3.2}). Subsequently, we leverage the constructed model capability to guide the fusion of 2D point features to obtain fused 2D point features (Sec. \ref{sec:3.3}). Finally, we distill a 3D MinkowskiNet from the fused 2D point features (Sec. \ref{sec:3.4}). Details are in Sec. \ref{sec:3.5}.

\subsection{Model Alignment via Text}
\label{sec:3.1}

\begin{figure}[t]
  \centering
  \includegraphics[scale=0.14]{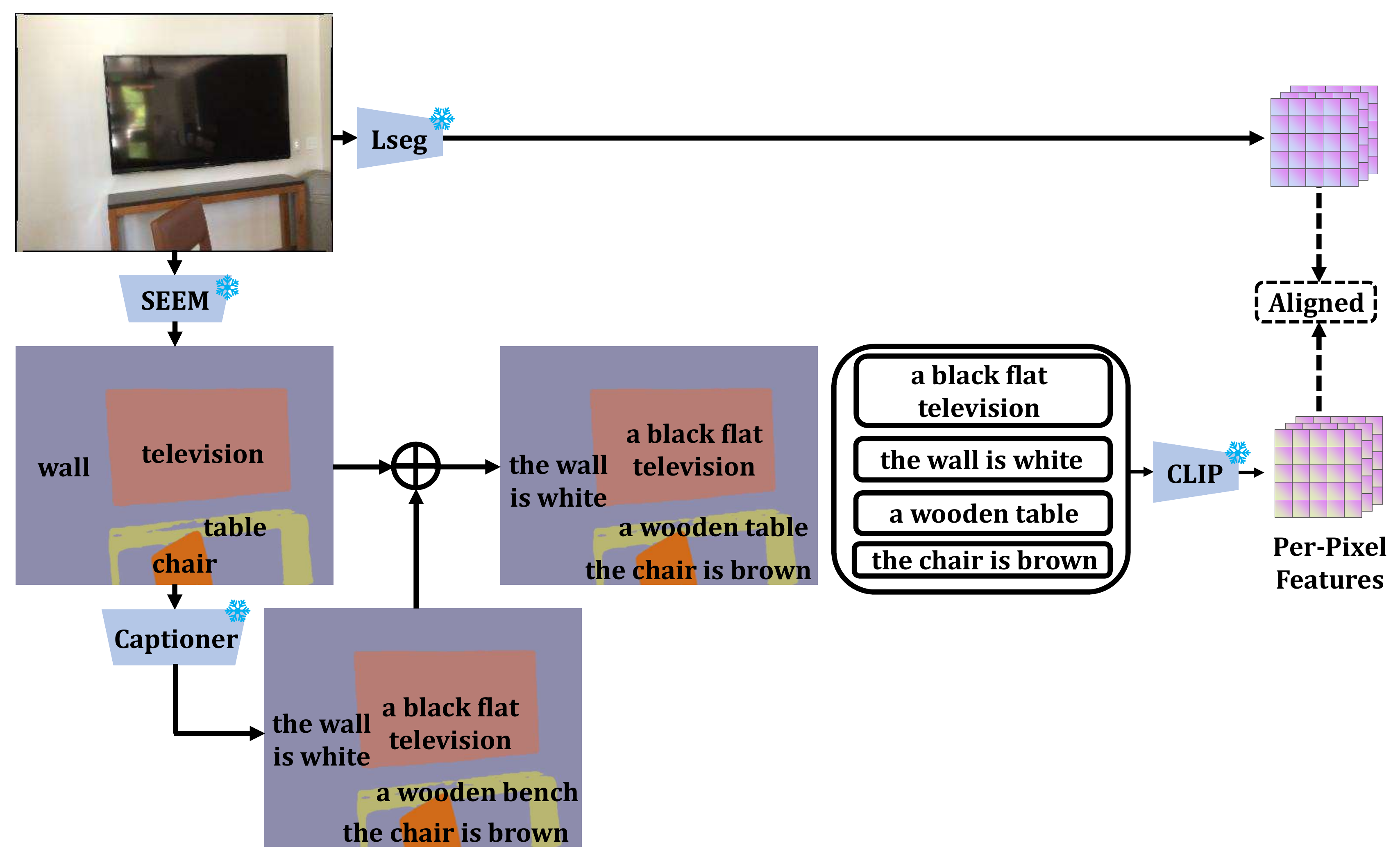}
  \caption{
  \textbf{Overview of Model Alignment via Text.} Features from different models are first aligned on text level, which are then encoded by a shared text encoder to produce aligned features.
  }
  \label{fig:align_model}
  \vspace{-10pt}
\end{figure}
\begin{figure*}[t]
  \centering
  \includegraphics[scale=0.45]{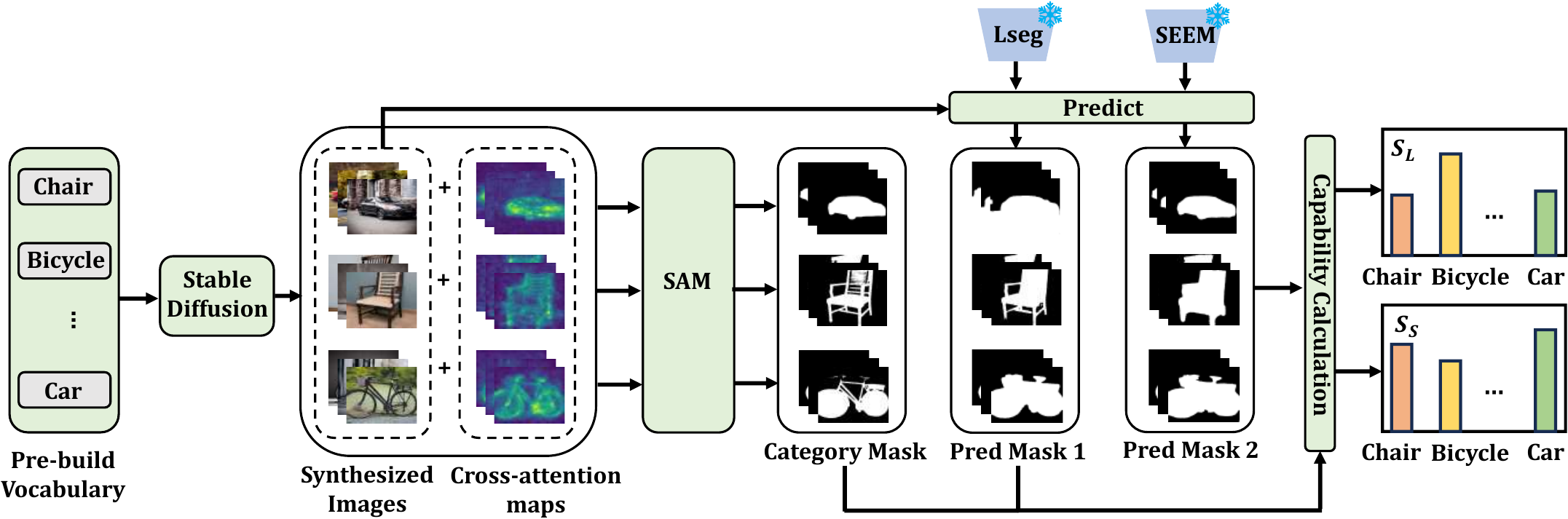}
  \caption{\textbf{Overview of Annotation-free Model Capability Construction.} Stable Diffusion model \cite{stablediffusion} is utilized to generate synthesized images with masks computed by SAM \cite{sam}. By assessing model's performance on synthesized images, we construct model capabilities.}
  \label{fig:capability}
  \vspace{-10pt}
\end{figure*}
In this section, our objective is to align two 2D models into the same feature space. Here, we take LSeg \cite{lseg} and SEEM \cite{SEEM} as an example without loss of generalizability. 
2D open vocabulary models usually consist of a text encoder and an image encoder that project text and images into a unified embedding space.  LSeg \cite{lseg} outputs dense per-pixel feature that is aligned with CLIP \cite{clip}. SEEM \cite{SEEM} is a maskformer \cite{mask2former} style model that predicts masks for an image and computes labels for each mask. However, SEEM \cite{SEEM}'s embedding space differs from LSeg’s due to variations in their text encoders. To address this, we align their embedding spaces. 

Fig. \ref{fig:align_model} shows an example of the aligning process. Given an image $I_i\in \mathbb{R}^{H\times W}$, we first input it into LSeg to get per-pixel features $f^{2D}_L\in \mathbb{R}^{H\times W \times C}$. Similarly, we input $I_i$ into SEEM to obtain the mask-label pairs. Specifically, we denote the $j$-th mask and its corresponding label as $m_{i,j}$ and $l_{i,j}$. Therefore, the output of SEEM can be formulated as $S_i=\{m_{i,j},l_{i,j}\}_{j=1,2,...,N_i}$, where $N_i$ is the total number of masks predicted in $I_i$. However, $l_{i,j}$ are simple nouns that lack intra-class diversity, e.g., color and shape. Therefore, we adopt a pre-trained captioner, TAP \cite{tap}, to generate captions $\{c_{i,j}\}_{j=1,2,...,N_i}$ for each mask $m_{i,j}$, providing additional semantic information. We then replace the noun labels in $\{c_{i,j}\}$ with the none labels predicted by SEEM $l_{i,j}$ to obatin $\{\hat{c}_{i,j}\}_{j=1,2,...,N_i}$, e.g., $c_{i,j}$=``a woodeow bench", $l_{i,j}$=``table" and $\hat{c}_{i,j}$=``a wooden table". Now we have obtained semantically-rich captions $\hat{c}_{i,j}$ for each mask, which can then be encoded by CLIP and mapped back to the image to form per-pixel features $f^{2D}_S\in \mathbb{R}^{H\times W \times C}$, which are aligned with $f^{2D}_L$ in CLIP's feature space.

We then follow OpenScene \cite{openscene} to calculate the pixel-point correspondence, which is then utilized to map the pixel features onto point features with a multi-view fusion strategy \cite{openscene}. The point features from LSeg and SEEM are denoted as $F^{2D}_L\in \mathbb{R}^{N\times C}$ and $F^{2D}_S\in \mathbb{R}^{N\times C}$, where $N$ is the number of points in the point cloud.

\subsection{Annotation-free Model Capability Construction}
\label{sec:3.2}

In this section, we aim to quantify model capabilities. However, obtaining test images and their annotations is difficult, making it impractical to directly assess the model’s performance. To overcome this, we utilize the Stable Diffusion (SD) \cite{stablediffusion} model and SAM \cite{sam} to establish a reference for evaluating model capabilities.  

As shown in Fig. \ref{fig:capability}, given a pre-built vocabulary consisting of common classes in the scene, denoted as $C=\{C_1, C_2, ..., C_K\}$ where $K$ is the number of classes, we leverage the SD \cite{stablediffusion} model to generate $m$ images $\hat{I}_q=\{\hat{I}^q_1, \hat{I}^q_2,...,\hat{I}^q_m\}$ for each class, where $q$ indicates the $q$-th class and $\hat{I}^q_j$ is the $j$-th generated image of the $q$-th class.
In addition to synthesizing a picture of a certain class, we need to localize the corresponding object in it with the help of cross-attention maps inside SD \cite{stablediffusion} models. Considering a conditional diffusion model (e.g., SD \cite{stablediffusion} model), the input to a UNet layer are noisy image features $F_v\in \mathbb{R}^{H\times W\times C}$ and text prompt features $F_p\in \mathbb{R}^{N\times D}$. The noisy image features are squeezed and then projected into the query $Q$ as $Q=F_cW^Q$, while text prompt features are projected into the key $K$ and the value $V$ similarly as $K=F_pW^K$ and $V=F_pW^V$, where $W^Q,W^K, W^V$ are the matrices of corresponding linear layers. Following this, the cross-attention map $M$ can be calculated as:
\begin{align}
\label{eq.1}
    M = \mathbf{Softmax}(\frac{QK^T}{\sqrt{D}}).
\end{align}
We denote $M_x^{y,z}$ as the cross-attention map of the $x$-th token in the $y$-th UNet layer at diffusion step $z$. To obtain more robust cross-attention maps, we follow \cite{diffumask} to aggregate all attention maps at all $Z$ time steps in all $Y$ UNet layes as:
\begin{equation}
\label{eq.2}
    \Bar{M}_x=\frac{1}{Y\cdot Z} \sum_{y\in Y, z\in Z} \frac{M_{x}^{y,z}}{\mathbf{max}(M_{x}^{y,z})}.
\end{equation}

We can now obtain a coarse mask for the target object in the image by binarizing the cross-attention map $ \Bar{M}_x$ via thresholding. Further, we leverage SAM \cite{sam} to generate precise mask predictions. Specifically, we sample several points within the coarse mask as the point prompt for the SAM model \cite{sam} to generate accurate masks $\mathbf{M}^{Pseudo}_{i,j}$, where $i$ and $j$ denote the $i$-th image of the $j$-th class. Similarly, LSeg and SEEM can be applied to the synthesized images to obtain masks $\mathbf{M}^{LSeg}_{i,j}$ and $\mathbf{M}^{SEEM}_{i,j}$ respectively. 

Following this, the metric mean intersection over union (mIOU) is adopted to measure model's capability. Specifically, the capability of LSeg \cite{lseg} model for a certain class $C_j$ can be constructed as :
\begin{equation}
\label{eq.3}
    S^{LSeg}_j=\frac{1}{m} \sum_{i=1,...,m} {\mathbf{mIoU}(\mathbf{M}^{Pseudo}_{i,j}, \mathbf{M}^{LSeg}_{i,j})},
\end{equation}
where $S^{LSeg}_j$ is the model capability of LSeg for the $j$-th class. $\mathbf{mIoU}(a, b)$ computes the mIoU between two binary masks. Similarly, the capability of SEEM model for a certain class $C_j$ can be constructed as:
\begin{equation}
\label{eq.4}
    S^{SEEM}_j=\frac{1}{m} \sum_{i=1,...,m} {\mathbf{mIoU}(\mathbf{M}^{Pseudo}_{i,j}, \mathbf{M}^{SEEM}_{i,j})}.
\end{equation}
Follwoing Eq. \ref{eq.3} and Eq. \ref{eq.4}, model capabilities for LSeg \cite{lseg} and SEEM \cite{SEEM} can be established as $S_L$ and $S_S$:

\begin{equation}
\begin{aligned}
\label{eq.5}
    S_{L}&=[{S_1^{LSeg}},...,{S_K^{LSeg}}],\\
    S_{S}&=[{S_1^{SEEM}},...,{S_K^{SEEM}}].
\end{aligned}
\end{equation}

\subsection{Feature Fusion}
\label{sec:3.3}

Now we have obtained aligned point features $F^{2D}_L$ and $F^{2D}_S$ and the corresponding constructed capability $S_{L}$ and $S_{S}$. In this section, our objective is to fuse the point features with the guide of model capabilities. 
We first encode the pre-built vocabulary $C=\{C_1,...,C_K\}$ using CLIP \cite{clip} to obtain text features ${F}_{text}=\{f_1, ..., f_K\}$. Next, we calculate the predicted category of each point for $F^{2D}_L$ and $F^{2D}_S$:

\begin{equation}
\begin{aligned}
\label{eq.6}
    \mathcal{P}_{LSeg}&=\underset{K}{\mathbf{argmax}}(F^{2D}_L\cdot F_{text}^{\mathbf{T}}),\\
    \mathcal{P}_{SEEM}&=\underset{K}{\mathbf{argmax}}(F^{2D}_S\cdot F_{text}^{\mathbf{T}}).
\end{aligned}
\end{equation}

We assume the right prediciton comes from either $\mathcal{P}_{LSeg}$ or $\mathcal{P}_{SEEM}$. Hence, we adopt the sum of the model capabilities of identifying $\mathcal{P}_{LSeg}$ and $\mathcal{P}_{SEEM}$ to measure the probability of the model making correct predictions. Specifically, the probability of LSeg \cite{lseg} and SEEM \cite{SEEM} making the right prediction can be formulated as:
\begin{equation}
\begin{aligned}
\label{eq.7}
    \mathcal{P}_{LSeg}&=\frac{S_L[\mathcal{P}_{LSeg}] + S_L[\mathcal{P}_{SEEM}]}{2},\\
    \mathcal{P}_{SEEM}&=\frac{S_S[\mathcal{P}_{LSeg}] + S_S[\mathcal{P}_{SEEM}]}{2}.
\end{aligned}
\end{equation}
Following this, the probability $\mathcal{P}_{LSeg}$ and $\mathcal{P}_{SEEM}$ can be the weights guiding the fusion of $F^{2D}_L$ and $F^{2D}_S$:
\begin{equation}
\begin{aligned}
\label{eq.8}
    {F}^{2D}_{fusion}=&\frac{\mathbf{exp}(\mathcal{P}_{LSeg}/\tau)}{\mathbf{exp}(\mathcal{P}_{LSeg}/\tau)+\mathbf{exp}(\mathcal{P}_{SEEM}/\tau)}F^{2D}_L+\\
    &\frac{\mathbf{exp}(\mathcal{P}_{SEEM}/\tau)}{\mathbf{exp}(\mathcal{P}_{LSeg}/\tau)+\mathbf{exp}(\mathcal{P}_{SEEM}/\tau)}F^{2D}_S.\\
\end{aligned}
\end{equation}
where $\tau$ is the temperature coefficient.

\subsection{Distillation}
\label{sec:3.4}

\label{Distillation}

\paragraph{Superpoint Distillation}
As 2D models could make potentially inconsistent predictions as shown in Fig. \ref{fig:intro}, we further introduce superpoint distillation to alleviate this problem, similar to GGSD \cite{GGSD}.

Specifically, for a given point cloud $\mathbf{P}\in \mathbb{R}^{N\times 3}$ and a point cloud encoder $\phi$, the encoded point features $F^{3D}$ is:
\begin{equation}
\begin{aligned}
\label{eq.9}
    F^{3D}=\phi(\mathbf{P}), \mathbb{R}^{N\times 3}\mapsto \mathbb{R}^{N\times C},
\end{aligned}
\end{equation}
where $N$ is the number of points and $C$ is the feature dimension.
We extract $L$ non-overlapping superpoints from $\mathbf{P}$ as $\{p_1,p_2,...,p_L\}$. We assume superpoints are semantically coherent that each superpoint should have the same category. The mean feature for each superpoint for $F^{2D}_{fusion}$ and $F^{3D}$ can be computed as: $f^{2D}_{SP}=\{\hat{f}^{2D}_1, \hat{f}^{2D}_2,...,\hat{f}^{2D}_L\}$ and $f^{3D}_{SP}=\{\hat{f}^{3D}_1, \hat{f}^{3D}_2,...,\hat{f}^{3D}_L\}$ respectively. Then we map the superpoint features $f^{2D}_{SP}$ onto all points to obtain per-point features $F^{2D}_{SP}$. The distillation is performed on both point level and superpoint level as:
\begin{equation}
\begin{aligned}
\label{eq.10}
    \mathcal{L}_p &= 1 - \mathbf{cos}(F^{2D}_{SP}, F^{3D}),\\
    \mathcal{L}_{sp} &= 1 - \mathbf{cos}(f^{2D}_{SP}, f^{3D}_{SP}),\\
    \mathcal{L} &= \mathcal{L}_p + \mathcal{L}_{sp},
\end{aligned}
\end{equation}
where $\mathbf{cos}$ computes the cosine similarity, $\mathcal{L}_p$ and $\mathcal{L}_{sp}$ indicate the distillation on point and superpoint level, $\mathcal{L}$ is the total loss. 
\vspace{-10pt}
\paragraph{Temporal Ensembling Self-Distillation}
The improvement in the distilled model inspires us to further exploit the model's potential via self-distillation. GGSD \cite{GGSD} proposes to use the student model's output to supervise the teacher model while updating the student model through EMA of the teacher model.
However, we find training the teacher model with a trainable and variable student model leads to an unstable training process, even a model collapse. Therefore, we propose temporal ensembling self-distillation.

Specifically, for a 3D network $\phi$ with its output $F^{3D}$, we construct $\hat{F}^{3D}$ to store $\phi$'s output in previous epochs to establish a more smooth optimization process: 
\begin{equation}
\begin{aligned}
\label{eq.11}
    \hat{F}^{3D}=\alpha \hat{F}^{3D} + (1-\alpha){F}^{3D},
\end{aligned}
\end{equation}
where $\alpha$ is a constant. By applying average pooling to superpoints of $\hat{F}^{3D}$, the pooled features $\hat{f}^{3D}_{SP}$ is then obtained. The pseudo label for $\hat{F}^{3D}$ and $\hat{f}^{3D}_{SP}$ can then be computed:
\begin{equation}
\begin{aligned}
\label{eq.12}
    \hat{\mathcal{P}}^{3D}&=\mathbf{argmax}(\hat{F}^{3D} \cdot \hat{F}_{text}^T),\\
    \hat{\mathcal{P}}^{3D}_{SP}&=\mathbf{argmax}(\hat{f}^{3D}_{SP} \cdot \hat{F}_{text}^T),
\end{aligned}
\end{equation}
where $\hat{F}_{text}$ is the class embedding of the dataset. $\hat{\mathcal{P}}^{3D}$ and $\hat{\mathcal{P}}^{3D}_{SP}$ then supervise $\phi$ by optimizing the losses:
\begin{equation}
\begin{aligned}
\label{eq.13}
    \mathcal{L}_p^{ST} &= \mathbf{CE}(\mathbf{argmax}(F^{3D}\cdot \hat{F}_{text}^T),\hat{\mathcal{P}}^{3D}),\\
    \mathcal{L}_{sp}^{ST} &= \mathbf{CE}(\mathbf{argmax}(f^{3D}_{SP}\cdot \hat{F}_{text}^T), \hat{\mathcal{P}}^{3D}_{SP}),\\
    \mathcal{L} &= \mathcal{L}_p^{ST} + \mathcal{L}_{sp}^{ST},
\end{aligned}
\end{equation}
where $\mathbf{CE}$ denotes the CrossEntropy loss, $\mathcal{L}_p^{ST}$ and $\mathcal{L}_{sp}^{ST}$ indicate the self distillation on point level and superpoint level respectively and $\mathcal{L}$ is the total loss.

\subsection{Training and Inference}
\label{sec:3.5}

\paragraph{Training}
The training process contains 100 epochs. In the first 70 epochs, we employ superpoint distillation. In the last 30 epochs, we employ both superpoint distillation and temporal ensembling self-distillation. Additionally, we designed two pre-build vocabularies for indoor scenes and outdoor scenes respectively. The detail is included in supplementary materials. While outdoor point clouds (e.g., nuScenes \cite{nuscenes}) are typically dominated by “roads”, we do not compute superpoints for nuScenes. Instead, we treat every single point in nuScenes as a superpoint for simplicity.

\paragraph{Inference}
We directly use the output features from the 3D model to calculate the similarity with CLIP \cite{clip} features of different categories without any other post-processing, e.g., superpoint or 2D-3D ensemble \cite{openscene}.

\section{Experiments}
We conduct extensive experiments to demonstrate the effectiveness of \name{} on 3D scene understanding tasks in a \textbf{zero-shot} fashion. We first introduce the experiment setup in Sec. \ref{sec:4.1}. In Sec. \ref{sec:4.2}, We evaluate \name{} on zero-shot open vocabulary semantic segmentation tasks. We then perform comprehensive ablation studies to validate our design choices in Sec. \ref{sec:4.3}. Further, we extend \name{} to instance segmentation and gaussian segmentation in Sec. \ref{sec:4.4}.

\subsection{Experiment Setup}
\label{sec:4.1}
\paragraph{Dataset.}
We evaluate \name{} on three commonly used datasets: ScanNet v2 \cite{scannet}, Matterport3D \cite{matterport3d} and nuScenes \cite{nuscenes}. 
ScanNet v2 \cite{scannet} is an indoor dataset, containing 1513 room scans from 2.5 million RGB-D images, the
average point number of which is 148k 
Matterport3D \cite{matterport3d} is another complex indoor dataset consisting of 10800 scenes of the building environment, equipped with 194K RGB-D images.
nuScenes \cite{nuscenes} is an outdoor dataset containing 1,000 driving sequences in total with 34k LiDAR point clouds.
ScanNet v2 \cite{scannet}, Matterport3D \cite{matterport3d}, and nuScenes \cite{nuscenes} follow the official data splits to generate corresponding training set, validation set and test set, with evaluations on 20, 21, and 23 categories respectively. ScanNet v2 \cite{scannet} validation set, Matterport3D \cite{matterport3d} test set, and, nuScenes \cite{nuscenes} validation set are adopted for evaluation.
\vspace{-10pt}

\paragraph{Implementation Details.}
We follow the settings specified in OpenScene \cite{openscene}, utilizing MinkowskiNet18A \cite{minknet} as the 3D backbone network, and adopting a voxel size of 2cm and 5cm for indoor datasets and ourdoor datasets respectively. Similarly with OpenScene \cite{openscene}, we employ a prompt engineering that modifies each class name "XX" to "a XX in a scene". AdamW \cite{adamw} is adopted for optimization. We use a batch size of 12 for ScanNet v2 \cite{scannet} and Matterport3D \cite{matterport3d} with 4 RTX 3090 GPUs. For nuScenes \cite{nuscenes}, we use a batch size of 8 with 4 A6000 GPUs. More details can be found in the supplementary file. Besides, based on the ablation table of OpenScene \cite{openscene}, we employ LSeg + SEEM for indoor datasets and OpenSeg + SEEM for outdoor dataset. Notably, OpenScene \cite{openscene} adopts a 2D-3D ensemble strategy to further enhance the performance, while we directly report the performance of the pure 3D model.

\subsection{Main Results}
\label{sec:4.2}
\paragraph{Evaluation on zero-shot 3D semantic segmentation.}
\begin{table}[t]
\caption{\textbf{Evaluations on zero-shot 3D semantic segmentation.} We compare \name{} with both zero-shot and fully-supervised approaches on nuScenes, ScanNet and MatterPort3D using mIOU as metrics. Best results under each setting are shown bold. }
\label{tab:main_result}
\vspace{-5pt}
\centering
\resizebox{0.45\textwidth}{!}{
\begin{tabular}{l|c|c|c}
    \toprule
    {} & \multicolumn{1}{c|}{nuScenes} & \multicolumn{1}{c|}{ScanNet v2} & \multicolumn{1}{c}{Matterport3D}\\
    \midrule
    \multicolumn{4}{l}{\textit{Fully-supervised approaches}} \\
    PointNet++~\cite{pointnet++}  & -  &{53.5} & -\\
    PointConv~\cite{pointconv}   & -  &{61.0} & -\\
    KPConv~\cite{kpconv} & -  &{69.2} & -\\
    Mix3D~\cite{mix3d} & -  &{73.6} & -\\
    LidarMultiNet~\cite{lidarmultinet} & \textbf{82.0}  &- & -\\
    PTv3~\cite{PTv3}  & 80.4  &\textbf{77.5} & -\\
    MinkowskiNet~\cite{minknet} & 78.0  & 69.0 & 54.2 \\ \hline
    \multicolumn{4}{l}{\textit{Zero-shot approaches}} \\
    OpenScene~\cite{openscene} & 42.1  & 54.2  & 43.4  \\
    GGSD~\cite{GGSD} & 46.1  & 56.5  & 40.1  \\
    Diff2Scene~\cite{diff2scene} &-  & 48.6  & 45.5  \\
    Seal~\cite{seal} & 45.0  &-  &-  \\
    OV3D~\cite{ov3d} & 44.6  & 57.3  & 45.8  \\
    \textbf{Ours}  & \textbf{47.5}  & \textbf{61.9}  & \textbf{48.6} \\ 
    \bottomrule
\end{tabular}
}
\vspace{-12pt}
\end{table}
Our proposed \name{} is compared with both fully-supervised and zero-shot methods on ScanNet v2 \cite{scannet}, Matterport3D \cite{matterport3d} and nuScenes \cite{nuscenes}. \name{} exhibits superior performance in zero-shot 3D semantic segmentation, as shown in Tab. \ref{tab:main_result}. Specifically, \name{} outperforms all previous methods, including the recent state-of-the-art OV3D \cite{ov3d}. Compared with previous SOTAs, \name{} shows a +1.4\%, +4.6\% and +2.8\% improvement in mIoU on nuScenes \cite{nuscenes}, ScanNet v2 \cite{scannet}and Matterport3D \cite{matterport3d} respectively.
\vspace{-10pt}

\paragraph{Evaluation in long-tail scenarios.}
Different from the standard benchmarks with dozens of categories, we further evaluate \name{} under more complex long-tail scenarios with more categories (e.g., 160) to validate its open vocabulary capability. The Matterport3D \cite{matterport3d} benchmark additionally provides K most common categories from the NYU label set for K = 40, 80, 160, which we evaluate on as shown in Tab. \ref{tab:long_tail}. Complying the zero-shot setting, we adopt the same distilled model to evaluate for all K = 40, 80, 160. The result suggests that our proposed \name{} consistently outperforms OpenScene \cite{openscene}, demonstrating its strong open vocabulary capability in long-tail scenarios.
\vspace{-10pt}

\begin{table}[t]
\label{tab:long_tail}
\caption{\textbf{Evaluation in long-tail scenarios.} We evaluate SAO in MatterPort40, MatterPort80 and MatterPort160 using mIOU as the metric. Best results are shown bold. }
\label{tab:long_tail}
\vspace{-5pt}
\centering
\resizebox{0.48\textwidth}{!}{
\begin{tabular}{l|c|c|c|c}
    \toprule
    {} & $K=21$ & $K=40$ & $K=80$ & $K=160$\\
    \midrule
    OpenScene~\cite{openscene} & 43.4 & 22.9&11.3&5.8\\ \hline
    \textbf{Ours} &\textbf{48.6} & \textbf{28.5} & \textbf{16.0} & \textbf{8.1}\\
    \bottomrule
\end{tabular}
}
\vspace{-1.4em}
\end{table}

\begin{figure*}[t]
  \centering
  \includegraphics[scale=0.18]{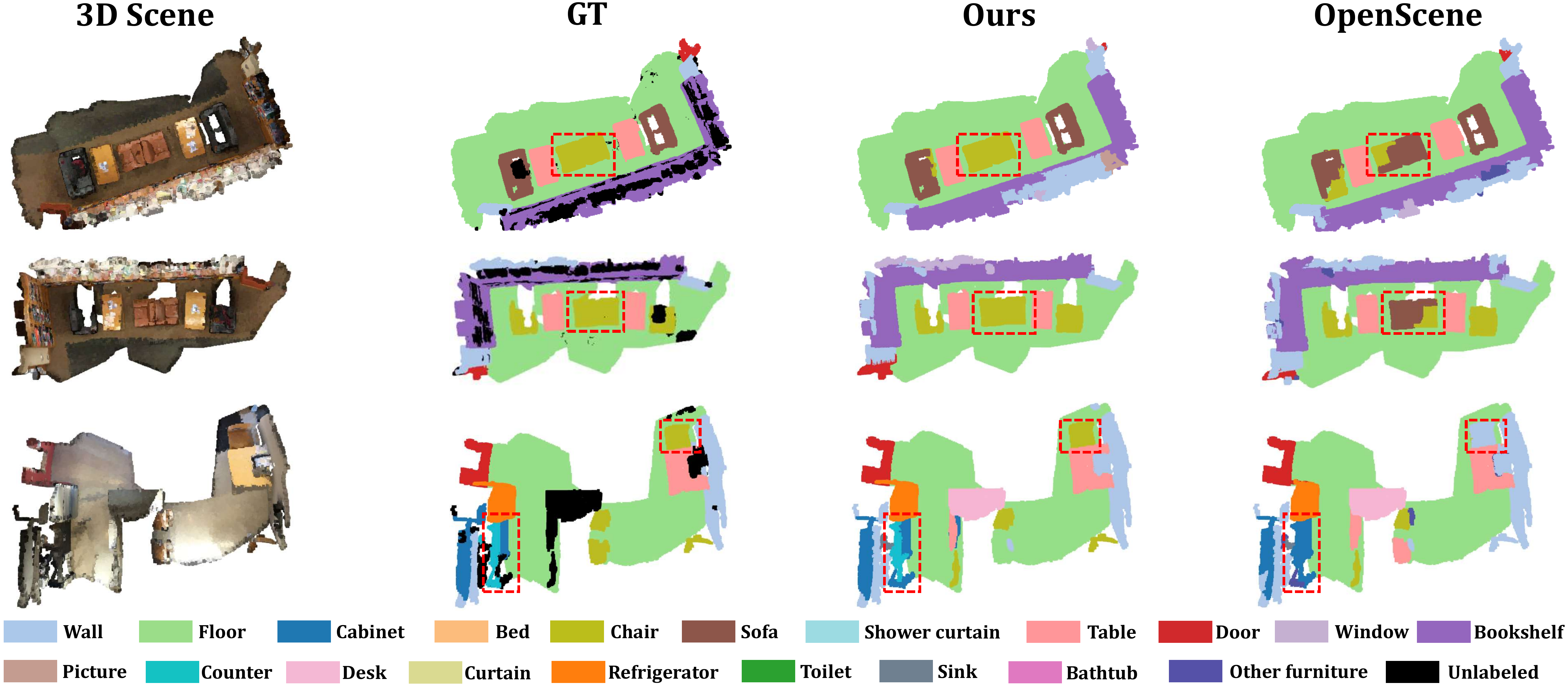}
  \caption{\textbf{Visualization results.} Semantic segmentation results of \name{} on ScanNet v2.}
  \label{fig:visualiza_scannet}
  \vspace{-10pt}
\end{figure*}

\paragraph{Visual Comparison}
Visual comparisons with OpenScene \cite{openscene} on semantic segmentation in ScanNet v2 \cite{scannet} are shown in Fig. \ref{fig:visualiza_scannet}, which shows that our approach effectively corrects some wrong predictions made by OpenScene \cite{openscene}. For example, OpenScene \cite{openscene} misidentifies a chair as a sofa, while our approach can easily tell the difference between a chair and a sofa. Additional visualization about querying different objects in a scene is provided in Fig. A in supplementary material. Visualization about performance on other datasets is provided in Fig. B, Fig. C and Fig. D.

\subsection{Ablation Studies and Analysis}
\label{sec:4.3}

\paragraph{Different 2D Features.} 
Top three lines of Tab. \ref{tab:ablation} show the performance obtained by directly projecting different 2D features onto 3D points. Based on the findings from OpenScene \cite{openscene}, we employ Lseg + SEEM for indoor scenes and OpenSeg + SEEM for outdoor scenes. Tab. \ref{tab:ablation} suggests that the performance of these models on different datasets varies randomly, e.g., SEEM \cite{SEEM} performs better than LSeg \cite{lseg} on Matterport3D \cite{matterport3d} (+1.6\%) while SEEM \cite{SEEM} performs worse than LSeg \cite{lseg} (-3.9\%) on ScanNet v2 \cite{scannet}. However, with our proposed fusion strategy, the performance after fusion is always better than that of two separate models. 
\vspace{-10pt}

\begin{table}[t]
\caption{\textbf{Ablation study of \name{}.} Zero-shot semantic segmentation performance of certain component of SAO being ablated. SP, TESD and Dis. indicates superpoint, temporal ensembling self distillation and distillation respectively.}
\label{tab:ablation}
\vspace{-5pt}
\centering
\resizebox{0.48\textwidth}{!}{
\begin{tabular}{l|c|c|c}
    \toprule

    {}  & {ScanNet v2}  & {Matterport}  & {nuScenes}\\
    \midrule

    2D Fusion LSeg~\cite{lseg}  & 51.2  & 38.6 &-\\
    2D Fusion OpenSeg~\cite{openseg}  & -  & - & 32.9\\
    2D Fusion SEEM~\cite{SEEM}  & 47.3  & 40.2 & 37.8\\
    2D Fusion Add   & 48.6  & 39.1 & 34.1\\
    2D Fusion Linear   & 49.9  & 39.4 & 34.8\\
    2D Fusion (Ours)  & \textbf{55.5}  & \textbf{43.6} & \textbf{40.0}\\
    \midrule
    3D Pixel-Point Dis.  & 56.7  & 45.1 & 45.4\\
    3D SP Dis.    & 59.2  & 46.3 & 45.4\\
    3D SP Dis. + TESD (Ours)   & \textbf{61.9}  & \textbf{48.6} & \textbf{47.5}\\
    \bottomrule
\end{tabular}

}
\vspace{-10pt}
\end{table}

\paragraph{Fusion strategy.} 
Tab. \ref{tab:ablation} also performs ablations of different fusion strategies including adding, linear fusion and our proposed fusion strategy in Sec. \ref{sec:3.3}. 
Adding indicates adding two features directly. Linear fusion computes the similarity between a point feature and text features and selects the largest similarity as the point feature's weight, based on which two point features are fused.
As can be seen, among all datasets, our proposed strategy performs the best. This indicates simple summation or linear fusion fails to inject model capabilities as guides into the feature fusion process, while our approach solves this.
\vspace{-10pt}

\paragraph{Distillation Pattern.} 
Bottom half of Tab. \ref{tab:ablation} shows the performance of the model distilled from 2D fused faetures in different ways, including pixel-point distillation, superpoint distillation, and temporal ensembling self distillation. OpenScene \cite{openscene} adopts pixel-point distillation to directly align point features and text features.
Note that the superpoint distillation performs the same with pixel-point distillation on nuScenes, because we treat every point in nuScenes as a superpoint in Sec. \ref{sec:3.5}.
The pixel-point distillation has already exhibited better performance than 2D features without additional designs, resulting in a +1.2\%, +1.5\%, and +5.4\% increase in mIoU on ScanNet v2 \cite{scannet}, Matterport3D \cite{matterport3d} and nuScenes \cite{nuscenes} respectively. Superpoint distillation brings extra inherent structural information of point clouds into distillation and achieves more accurate predictions. Temporal ensembling self distillation further exploits the potential of distilled models via self distillation, boosting the performance gap with 2D features to +6.4\%, +5.0\%, and +7.5\% in mIoU.

\begin{table}[!ht]
\caption{\textbf{Fusion of LSeg, SEEM and ODISE.} Dis. indicates distillation.}
\label{tab:triple}
\vspace{-5pt}
\centering
\resizebox{0.48\textwidth}{!}{
\begin{tabular}{l|c|c}
    \toprule

    {}  & {ScanNet v2}  & {Matterport3D} \\
    \midrule

    2D Fusion LSeg~\cite{lseg}  & 51.2  & 43.5 \\
    2D Fusion SEEM~\cite{SEEM}  & \textbf{47.3}  & 46.1 \\
    2D Fusion ODISE~\cite{odise}  & 46.9  & \textbf{46.7} \\

    \midrule

    3D SEEM + LSeg Dis.  & 61.9  & 48.6 \\
    3D SEEM + LSeg + ODISE Dis.  & \textbf{62.5}  & \textbf{49.8} \\

    \bottomrule
\end{tabular}
}
\vspace{-10pt}
\end{table}

\vspace{-10pt}

\paragraph{More 2D Models.}
We further evaluate the effectiveness of \name{} by incorporating additional 2D open-vocabulary models. As detailed in Table~\ref{tab:triple}, we integrate another 2D model, ODISE \cite{odise}, alongside LSeg \cite{lseg} and SEEM \cite{SEEM}, conducting experiments on the ScanNet v2 and Matterport3D. The results demonstrate that the inclusion of ODISE leads to a further performance enhancement, while also highlighting the promising scalability of our approach.

\subsection{Further Exploration}
\label{sec:4.4}
We further apply \name{} to other tasks, including 3D Gaussian segmentation and 3D instance segmentation. 
\begin{table}[t]
\caption{\textbf{Gaussian segmentation}. 2D semantic segmentation results of 3D gaussian splatting on 12 scenes from the ScanNet v2 validation set. Comparison focuses on NeRF/3DGS-based methods using mIoU and mAcc.}
\label{tab:gaussian}
\vspace{-5pt}
\centering
\resizebox{0.48\textwidth}{!}{
\begin{tabular}{lccc}
\toprule
\textbf{Method} & \textbf{Backbone} & \textbf{mIoU} & \textbf{mAcc} \\
\midrule
\multicolumn{4}{l}{\textit{Fully-supervised approaches}} \\
Mask2Former~\cite{mask2former} & ViT~\cite{dosovitskiy2020image} & 46.7 & - \\
DM-NeRF~\cite{dmnerf} & NeRF~\cite{mildenhall2021nerf} & 49.5 & - \\
SemanticNeRF~\cite{semanticnerf} & NeRF & 59.2 & - \\
Panoptic Lifting~\cite{panopticlifting} & NeRF & \textbf{65.2} & - \\
\midrule
\multicolumn{4}{l}{\textit{Zero-shot approaches}} \\
OpenSeg~\cite{openseg} & EfficientNet~\cite{tan2019efficientnet} & 53.4 & 75.1 \\
LSeg~\cite{lseg} & ViT & 56.1 & 74.5 \\
LERF~\cite{lerf} & NeRF+CLIP~\cite{clip} & 31.2 & 61.7 \\
PVLFF~\cite{pvlff} & NeRF+LSeg & 52.9 & 67.0 \\
LangSplat~\cite{langsplat} & 3DGS~\cite{kerbl20233d}+CLIP & 24.7 & 42.0 \\
Feature3DGS~\cite{feature3dgs} & 3DGS+LSeg & 59.2 & 75.1 \\
Semantic Gaussians~\cite{guo2024semantic} & 3DGS+LSeg & 60.7 & 76.3 \\
Ours & 3DGS+LSeg+SEEM~\cite{zou2024segment} & \textbf{63.9} & \textbf{79.9} \\
\bottomrule
\end{tabular}
}
\vspace{-10pt}
\end{table}
\vspace{-10pt}
\paragraph{3D Gaussian Segmentation}
3D Gaussian segmentation aims to assign a semantic label to each Gaussian point, facilitating the rendering of arbitrary 2D views for a thorough understanding of 3D scenes. In this task, we use Semantic Gaussians as our baseline, harnessing the 2D foundation model LSeg to achieve zero-shot 3D Gaussian segmentation via knowledge distillation. As shown in Tab.~\ref{tab:gaussian}, rather than relying exclusively on LSeg, we integrate the strengths of both LSeg and SEEM to improve scene understanding, achieving a performance increase of 3.2 in mIoU and 3.6 in mAcc. Notably, due to the distinct 3D representations (Gaussian vs point cloud), we redistill a new 3D encoder tailored specifically for 3D Gaussian. During this process, we do not utilize the distillation strategy outlined in Sec.~\ref{Distillation}. This task further highlights the effectiveness and generalizability of our approach.
\vspace{-10pt}

\paragraph{3D Instance Segmentation}

3D instance segmentation seeks to detect and delineate multiple instances of specific object categories within a 3D space. In this task, building on previous approaches, we employ Mask3D \cite{schult2022mask3d} to generate mask proposals and then leverage the distilled point encoder, as presented in Tab.~\ref{tab:main_result}, to derive the corresponding semantics. As demonstrated in Tab.~\ref{tab:ins}, our method achieves significant improvements over the baseline OpenScene, with an mAP of 5.3, AP@50 of 6.0, and AP@25 of 6.3. Compared to the current state-of-the-art OpenIns3D, our approach boosts mAP by 5.2 and AP@50 by 2.8.
\begin{table}[t]
\caption{\textbf{Instance segmentation}. 3D open-vocabulary instance segmentation results on ScanNet v2 validation set.}
\label{tab:ins}
\vspace{-5pt}
\centering
\resizebox{0.48\textwidth}{!}{
\begin{tabular}{lcccc}
\toprule
\textbf{Method}&\textbf{Semantic}&\textbf{mAP}&\textbf{AP@50} & \textbf{AP@25} \\
\midrule
PointClip~\cite{zhang2022pointclip} & Clip~\cite{radford2021learning}&  -& 4.5 &14.4\\
OpenIns3D~\cite{huang2024openins3d} &Yoloworld~\cite{cheng2024yolo}&19.9&28.9& \textbf{38.9}\\
OpenScene~\cite{openscene} (2D/3D Ens.)& LSeg~\cite{lseg}&19.7&25.9 &30.4 \\
OpenScene (3D Distill)& LSeg&19.8&25.7 &30.4 \\
Ours&LSeg+SEEM~\cite{zou2024segment}&\textbf{25.1}&\textbf{31.7}&36.7\\ 

\bottomrule
\end{tabular}
}
\vspace{-10pt}
\end{table}

\section{Conclusion}
In this paper, we present SAS, a simple yet effective approach to transfer the open vocabulary capabilities of multiple pre-trained 2D models to 3D domain. By aligning the embedding space of different 2D models and utilizing diffusion models to construct the model capability, which then guides the fusion of different features, SAS achieves superior performance on both indoor and outdoor datasets over previous methods. Additionally, SAS exhibits strong generalization by extending to other tasks, including gaussian segmentation and instance segmentation.

{\small
\bibliographystyle{ieeenat_fullname}
\bibliography{main}

\begin{thebibliography}{97}
\providecommand{\natexlab}[1]{#1}
\providecommand{\url}[1]{\texttt{#1}}
\expandafter\ifx\csname urlstyle\endcsname\relax
  \providecommand{\doi}[1]{doi: #1}\else
  \providecommand{\doi}{doi: \begingroup \urlstyle{rm}\Url}\fi

\bibitem[Caesar et~al.(2020)Caesar, Bankiti, Lang, Vora, Liong, Xu, Krishnan, Pan, Baldan, and Beijbom]{nuscenes}
Holger Caesar, Varun Bankiti, Alex~H Lang, Sourabh Vora, Venice~Erin Liong, Qiang Xu, Anush Krishnan, Yu Pan, Giancarlo Baldan, and Oscar Beijbom.
\newblock nuscenes: A multimodal dataset for autonomous driving.
\newblock In \emph{Proceedings of the IEEE/CVF conference on computer vision and pattern recognition}, pages 11621--11631, 2020.

\bibitem[Cen et~al.(2023)Cen, Zhou, Fang, Shen, Xie, Jiang, Zhang, Tian, et~al.]{cen2023segment}
Jiazhong Cen, Zanwei Zhou, Jiemin Fang, Wei Shen, Lingxi Xie, Dongsheng Jiang, Xiaopeng Zhang, Qi Tian, et~al.
\newblock Segment anything in 3d with nerfs.
\newblock \emph{Advances in Neural Information Processing Systems}, 36:\penalty0 25971--25990, 2023.

\bibitem[Chang et~al.(2017)Chang, Dai, Funkhouser, Halber, Niessner, Savva, Song, Zeng, and Zhang]{matterport3d}
Angel Chang, Angela Dai, Thomas Funkhouser, Maciej Halber, Matthias Niessner, Manolis Savva, Shuran Song, Andy Zeng, and Yinda Zhang.
\newblock Matterport3d: Learning from rgb-d data in indoor environments.
\newblock \emph{arXiv preprint arXiv:1709.06158}, 2017.

\bibitem[Chen et~al.(2024)Chen, Blomqvist, Milano, and Siegwart]{pvlff}
Haoran Chen, Kenneth Blomqvist, Francesco Milano, and Roland Siegwart.
\newblock Panoptic vision-language feature fields.
\newblock \emph{IEEE Robotics and Automation Letters}, 2024.

\bibitem[Chen et~al.(2023)Chen, Liu, Kong, Zhu, Ma, Li, Hou, Qiao, and Wang]{clip2scene}
Runnan Chen, Youquan Liu, Lingdong Kong, Xinge Zhu, Yuexin Ma, Yikang Li, Yuenan Hou, Yu Qiao, and Wenping Wang.
\newblock Clip2scene: Towards label-efficient 3d scene understanding by clip.
\newblock In \emph{Proceedings of the IEEE/CVF Conference on Computer Vision and Pattern Recognition}, pages 7020--7030, 2023.

\bibitem[Cheng et~al.(2022)Cheng, Misra, Schwing, Kirillov, and Girdhar]{mask2former}
Bowen Cheng, Ishan Misra, Alexander~G Schwing, Alexander Kirillov, and Rohit Girdhar.
\newblock Masked-attention mask transformer for universal image segmentation.
\newblock In \emph{Proceedings of the IEEE/CVF conference on computer vision and pattern recognition}, pages 1290--1299, 2022.

\bibitem[Cheng et~al.(2024)Cheng, Song, Ge, Liu, Wang, and Shan]{cheng2024yolo}
Tianheng Cheng, Lin Song, Yixiao Ge, Wenyu Liu, Xinggang Wang, and Ying Shan.
\newblock Yolo-world: Real-time open-vocabulary object detection.
\newblock In \emph{Proceedings of the IEEE/CVF Conference on Computer Vision and Pattern Recognition}, pages 16901--16911, 2024.

\bibitem[Choy et~al.(2019)Choy, Gwak, and Savarese]{minknet}
Christopher Choy, JunYoung Gwak, and Silvio Savarese.
\newblock 4d spatio-temporal convnets: Minkowski convolutional neural networks.
\newblock In \emph{IEEE Conf. Comput. Vis. Pattern Recog.}, pages 3075--3084, 2019.

\bibitem[Dai et~al.(2017)Dai, Chang, Savva, Halber, Funkhouser, and Nie{\ss}ner]{scannet}
Angela Dai, Angel~X Chang, Manolis Savva, Maciej Halber, Thomas Funkhouser, and Matthias Nie{\ss}ner.
\newblock Scannet: Richly-annotated 3d reconstructions of indoor scenes.
\newblock In \emph{IEEE Conf. Comput. Vis. Pattern Recog.}, pages 5828--5839, 2017.

\bibitem[Deng et~al.(2023)Deng, Wang, Lu, He, Zhang, Yu, and Zhang]{deng2023se}
Jiacheng Deng, Chuxin Wang, Jiahao Lu, Jianfeng He, Tianzhu Zhang, Jiyang Yu, and Zhe Zhang.
\newblock Se-ornet: Self-ensembling orientation-aware network for unsupervised point cloud shape correspondence.
\newblock In \emph{Proceedings of the IEEE/CVF Conference on Computer Vision and Pattern Recognition}, pages 5364--5373, 2023.

\bibitem[Deng et~al.(2024{\natexlab{a}})Deng, Lu, and Zhang]{deng2024diff3detr}
Jiacheng Deng, Jiahao Lu, and Tianzhu Zhang.
\newblock Diff3detr: Agent-based diffusion model for semi-supervised 3d object detection.
\newblock In \emph{European Conference on Computer Vision}, pages 57--73. Springer, 2024{\natexlab{a}}.

\bibitem[Deng et~al.(2024{\natexlab{b}})Deng, Lu, and Zhang]{deng2024unsupervised}
Jiacheng Deng, Jiahao Lu, and Tianzhu Zhang.
\newblock Unsupervised template-assisted point cloud shape correspondence network.
\newblock In \emph{Proceedings of the IEEE/CVF Conference on Computer Vision and Pattern Recognition}, pages 5250--5259, 2024{\natexlab{b}}.

\bibitem[Deng et~al.(2025)Deng, Lu, and Zhang]{deng2025quantity}
Jiacheng Deng, Jiahao Lu, and Tianzhu Zhang.
\newblock Quantity-quality enhanced self-training network for weakly supervised point cloud semantic segmentation.
\newblock \emph{IEEE Transactions on Pattern Analysis and Machine Intelligence}, 2025.

\bibitem[Deprelle et~al.(2019)Deprelle, Groueix, Fisher, Kim, Russell, and Aubry]{deprelle2019learning}
Theo Deprelle, Thibault Groueix, Matthew Fisher, Vladimir Kim, Bryan Russell, and Mathieu Aubry.
\newblock Learning elementary structures for 3d shape generation and matching.
\newblock \emph{Advances in Neural Information Processing Systems}, 32, 2019.

\bibitem[Ding et~al.(2023)Ding, Yang, Xue, Zhang, Bai, and Qi]{pla}
Runyu Ding, Jihan Yang, Chuhui Xue, Wenqing Zhang, Song Bai, and Xiaojuan Qi.
\newblock Pla: Language-driven open-vocabulary 3d scene understanding.
\newblock In \emph{IEEE Conf. Comput. Vis. Pattern Recog.}, pages 7010--7019, 2023.

\bibitem[Dosovitskiy et~al.(2020)Dosovitskiy, Beyer, Kolesnikov, Weissenborn, Zhai, Unterthiner, Dehghani, Minderer, Heigold, Gelly, et~al.]{dosovitskiy2020image}
Alexey Dosovitskiy, Lucas Beyer, Alexander Kolesnikov, Dirk Weissenborn, Xiaohua Zhai, Thomas Unterthiner, Mostafa Dehghani, Matthias Minderer, Georg Heigold, Sylvain Gelly, et~al.
\newblock An image is worth 16x16 words: Transformers for image recognition at scale.
\newblock \emph{arXiv preprint arXiv:2010.11929}, 2020.

\bibitem[Ettinger et~al.(2021)Ettinger, Cheng, Qi, Zhou, et~al.]{waymo}
Scott Ettinger, Shuyang Cheng, Charles~R Qi, Yin Zhou, et~al.
\newblock Large scale interactive motion forecasting for autonomous driving: The waymo open motion dataset.
\newblock In \emph{CVPR}, pages 9710--9719, 2021.

\bibitem[Felzenszwalb and Huttenlocher(2004)]{effiseg}
Pedro~F Felzenszwalb and Daniel~P Huttenlocher.
\newblock Efficient graph-based image segmentation.
\newblock \emph{International journal of computer vision}, 59:\penalty0 167--181, 2004.

\bibitem[Geiger et~al.(2013)Geiger, Lenz, Stiller, and Urtasun]{kitti}
Andreas Geiger, Philip Lenz, Christoph Stiller, and Raquel Urtasun.
\newblock Vision meets robotics: The kitti dataset.
\newblock \emph{The International Journal of Robotics Research}, 32\penalty0 (11):\penalty0 1231--1237, 2013.

\bibitem[Ghiasi et~al.(2022{\natexlab{a}})Ghiasi, Gu, Cui, and Lin]{ghiasi2022scaling}
Golnaz Ghiasi, Xiuye Gu, Yin Cui, and Tsung-Yi Lin.
\newblock Scaling open-vocabulary image segmentation with image-level labels.
\newblock In \emph{European Conference on Computer Vision}, pages 540--557. Springer, 2022{\natexlab{a}}.

\bibitem[Ghiasi et~al.(2022{\natexlab{b}})Ghiasi, Gu, Cui, and Lin]{openseg}
Golnaz Ghiasi, Xiuye Gu, Yin Cui, and Tsung-Yi Lin.
\newblock Scaling open-vocabulary image segmentation with image-level labels.
\newblock In \emph{Eur. Conf. Comput. Vis.}, pages 540--557. Springer, 2022{\natexlab{b}}.

\bibitem[Groueix et~al.(2018)Groueix, Fisher, Kim, Russell, and Aubry]{groueix20183d}
Thibault Groueix, Matthew Fisher, Vladimir~G Kim, Bryan~C Russell, and Mathieu Aubry.
\newblock 3d-coded: 3d correspondences by deep deformation.
\newblock In \emph{Proceedings of the european conference on computer vision (ECCV)}, pages 230--246, 2018.

\bibitem[Guo et~al.(2024)Guo, Ma, Fan, Liu, and Li]{guo2024semantic}
Jun Guo, Xiaojian Ma, Yue Fan, Huaping Liu, and Qing Li.
\newblock Semantic gaussians: Open-vocabulary scene understanding with 3d gaussian splatting.
\newblock \emph{arXiv preprint arXiv:2403.15624}, 2024.

\bibitem[Huang et~al.(2024)Huang, Wu, Chen, Zhao, Zhu, and Lasenby]{huang2024openins3d}
Zhening Huang, Xiaoyang Wu, Xi Chen, Hengshuang Zhao, Lei Zhu, and Joan Lasenby.
\newblock Openins3d: Snap and lookup for 3d open-vocabulary instance segmentation.
\newblock In \emph{European Conference on Computer Vision}, pages 169--185. Springer, 2024.

\bibitem[Jiang et~al.(2020)Jiang, Zhao, Shi, Liu, Fu, and Jia]{jiang2020pointgroup}
Li Jiang, Hengshuang Zhao, Shaoshuai Shi, Shu Liu, Chi-Wing Fu, and Jiaya Jia.
\newblock Pointgroup: Dual-set point grouping for 3d instance segmentation.
\newblock In \emph{Proceedings of the IEEE/CVF conference on computer vision and Pattern recognition}, pages 4867--4876, 2020.

\bibitem[Jiang et~al.(2024{\natexlab{a}})Jiang, Shi, and Schiele]{jiang2024open}
Li Jiang, Shaoshuai Shi, and Bernt Schiele.
\newblock Open-vocabulary 3d semantic segmentation with foundation models.
\newblock In \emph{Proceedings of the IEEE/CVF Conference on Computer Vision and Pattern Recognition}, pages 21284--21294, 2024{\natexlab{a}}.

\bibitem[Jiang et~al.(2024{\natexlab{b}})Jiang, Shi, and Schiele]{ov3d}
Li Jiang, Shaoshuai Shi, and Bernt Schiele.
\newblock Open-vocabulary 3d semantic segmentation with foundation models.
\newblock In \emph{Proceedings of the IEEE/CVF Conference on Computer Vision and Pattern Recognition}, pages 21284--21294, 2024{\natexlab{b}}.

\bibitem[Kerbl et~al.(2023)Kerbl, Kopanas, Leimk{\"u}hler, and Drettakis]{kerbl20233d}
Bernhard Kerbl, Georgios Kopanas, Thomas Leimk{\"u}hler, and George Drettakis.
\newblock 3d gaussian splatting for real-time radiance field rendering.
\newblock \emph{ACM Trans. Graph.}, 42\penalty0 (4):\penalty0 139--1, 2023.

\bibitem[Kerr et~al.(2023{\natexlab{a}})Kerr, Kim, Goldberg, Kanazawa, and Tancik]{kerr2023lerf}
Justin Kerr, Chung~Min Kim, Ken Goldberg, Angjoo Kanazawa, and Matthew Tancik.
\newblock Lerf: Language embedded radiance fields.
\newblock In \emph{Proceedings of the IEEE/CVF International Conference on Computer Vision}, pages 19729--19739, 2023{\natexlab{a}}.

\bibitem[Kerr et~al.(2023{\natexlab{b}})Kerr, Kim, Goldberg, Kanazawa, and Tancik]{lerf}
Justin Kerr, Chung~Min Kim, Ken Goldberg, Angjoo Kanazawa, and Matthew Tancik.
\newblock Lerf: Language embedded radiance fields.
\newblock In \emph{Int. Conf. Comput. Vis.}, pages 19729--19739, 2023{\natexlab{b}}.

\bibitem[Kingma and Ba(2014)]{adam}
Diederik~P Kingma and Jimmy Ba.
\newblock Adam: A method for stochastic optimization.
\newblock \emph{arXiv preprint arXiv:1412.6980}, 2014.

\bibitem[Kirillov et~al.(2023)Kirillov, Mintun, Ravi, Mao, Rolland, Gustafson, Xiao, Whitehead, Berg, Lo, et~al.]{sam}
Alexander Kirillov, Eric Mintun, Nikhila Ravi, Hanzi Mao, Chloe Rolland, Laura Gustafson, Tete Xiao, Spencer Whitehead, Alexander~C Berg, Wan-Yen Lo, et~al.
\newblock Segment anything.
\newblock \emph{arXiv preprint arXiv:2304.02643}, 2023.

\bibitem[Kolodiazhnyi et~al.(2024)Kolodiazhnyi, Vorontsova, Konushin, and Rukhovich]{kolodiazhnyi2024oneformer3d}
Maxim Kolodiazhnyi, Anna Vorontsova, Anton Konushin, and Danila Rukhovich.
\newblock Oneformer3d: One transformer for unified point cloud segmentation.
\newblock In \emph{Proceedings of the IEEE/CVF Conference on Computer Vision and Pattern Recognition}, pages 20943--20953, 2024.

\bibitem[Lai et~al.(2023)Lai, Yuan, Chu, Chen, Hu, and Jia]{lai2023mask}
Xin Lai, Yuhui Yuan, Ruihang Chu, Yukang Chen, Han Hu, and Jiaya Jia.
\newblock Mask-attention-free transformer for 3d instance segmentation.
\newblock In \emph{Proceedings of the IEEE/CVF International Conference on Computer Vision}, pages 3693--3703, 2023.

\bibitem[Landrieu and Simonovsky(2018)]{landrieu2018large}
Loic Landrieu and Martin Simonovsky.
\newblock Large-scale point cloud semantic segmentation with superpoint graphs.
\newblock In \emph{Proceedings of the IEEE conference on computer vision and pattern recognition}, pages 4558--4567, 2018.

\bibitem[Lang et~al.(2019)Lang, Vora, Caesar, Zhou, Yang, and Beijbom]{lang2019pointpillars}
Alex~H Lang, Sourabh Vora, Holger Caesar, Lubing Zhou, Jiong Yang, and Oscar Beijbom.
\newblock Pointpillars: Fast encoders for object detection from point clouds.
\newblock In \emph{Proceedings of the IEEE/CVF conference on computer vision and pattern recognition}, pages 12697--12705, 2019.

\bibitem[Lang et~al.(2021)Lang, Ginzburg, Avidan, and Raviv]{lang2021dpc}
Itai Lang, Dvir Ginzburg, Shai Avidan, and Dan Raviv.
\newblock Dpc: Unsupervised deep point correspondence via cross and self construction.
\newblock In \emph{2021 International Conference on 3D Vision (3DV)}, pages 1442--1451. IEEE, 2021.

\bibitem[Li et~al.(2022{\natexlab{a}})Li, Weinberger, Belongie, Koltun, and Ranftl]{li2022language}
Boyi Li, Kilian~Q Weinberger, Serge Belongie, Vladlen Koltun, and Ren{\'e} Ranftl.
\newblock Language-driven semantic segmentation.
\newblock \emph{arXiv preprint arXiv:2201.03546}, 2022{\natexlab{a}}.

\bibitem[Li et~al.(2022{\natexlab{b}})Li, Weinberger, Belongie, Koltun, and Ranftl]{lseg}
Boyi Li, Kilian~Q. Weinberger, Serge~J. Belongie, Vladlen Koltun, and Ren{\'{e}} Ranftl.
\newblock Language-driven semantic segmentation.
\newblock In \emph{Int. Conf. Learn. Represent.}, 2022{\natexlab{b}}.

\bibitem[Li et~al.(2023)Li, Dai, Han, and Ding]{li2023mseg3d}
Jiale Li, Hang Dai, Hao Han, and Yong Ding.
\newblock Mseg3d: Multi-modal 3d semantic segmentation for autonomous driving.
\newblock In \emph{Proceedings of the IEEE/CVF conference on computer vision and pattern recognition}, pages 21694--21704, 2023.

\bibitem[Li et~al.(2022{\natexlab{c}})Li, Yu, Meng, Caine, Ngiam, Peng, Shen, Lu, Zhou, Le, et~al.]{li2022deepfusion}
Yingwei Li, Adams~Wei Yu, Tianjian Meng, Ben Caine, Jiquan Ngiam, Daiyi Peng, Junyang Shen, Yifeng Lu, Denny Zhou, Quoc~V Le, et~al.
\newblock Deepfusion: Lidar-camera deep fusion for multi-modal 3d object detection.
\newblock In \emph{Proceedings of the IEEE/CVF conference on computer vision and pattern recognition}, pages 17182--17191, 2022{\natexlab{c}}.

\bibitem[Li et~al.(2024)Li, Ai, Lu, Wang, Deng, Chang, Liang, Yang, Zhang, and Zhang]{li2024mamba24}
Zhuoyuan Li, Yubo Ai, Jiahao Lu, ChuXin Wang, Jiacheng Deng, Hanzhi Chang, Yanzhe Liang, Wenfei Yang, Shifeng Zhang, and Tianzhu Zhang.
\newblock Mamba24/8d: Enhancing global interaction in point clouds via state space model.
\newblock \emph{arXiv preprint arXiv:2406.17442}, 2024.

\bibitem[Liang et~al.(2023)Liang, Wu, Dai, Li, Zhao, Zhang, Zhang, Vajda, and Marculescu]{liang2023open}
Feng Liang, Bichen Wu, Xiaoliang Dai, Kunpeng Li, Yinan Zhao, Hang Zhang, Peizhao Zhang, Peter Vajda, and Diana Marculescu.
\newblock Open-vocabulary semantic segmentation with mask-adapted clip.
\newblock In \emph{Proceedings of the IEEE/CVF Conference on Computer Vision and Pattern Recognition}, pages 7061--7070, 2023.

\bibitem[Liu et~al.(2023)Liu, Kong, Cen, Chen, Zhang, Pan, Chen, and Liu]{seal}
Youquan Liu, Lingdong Kong, Jun Cen, Runnan Chen, Wenwei Zhang, Liang Pan, Kai Chen, and Ziwei Liu.
\newblock Segment any point cloud sequences by distilling vision foundation models.
\newblock \emph{Advances in Neural Information Processing Systems}, 36:\penalty0 37193--37229, 2023.

\bibitem[Loshchilov and Hutter(2017)]{adamw}
Ilya Loshchilov and Frank Hutter.
\newblock Decoupled weight decay regularization.
\newblock \emph{arXiv preprint arXiv:1711.05101}, 2017.

\bibitem[Lu et~al.(2023)Lu, Deng, Wang, He, and Zhang]{lu2023query}
Jiahao Lu, Jiacheng Deng, Chuxin Wang, Jianfeng He, and Tianzhu Zhang.
\newblock Query refinement transformer for 3d instance segmentation.
\newblock In \emph{Proceedings of the IEEE/CVF International Conference on Computer Vision}, pages 18516--18526, 2023.

\bibitem[Lu et~al.(2025)Lu, Deng, and Zhang]{lu2025beyond}
Jiahao Lu, Jiacheng Deng, and Tianzhu Zhang.
\newblock Beyond the final layer: Hierarchical query fusion transformer with agent-interpolation initialization for 3d instance segmentation.
\newblock \emph{arXiv preprint arXiv:2502.04139}, 2025.

\bibitem[Mildenhall et~al.(2021)Mildenhall, Srinivasan, Tancik, Barron, Ramamoorthi, and Ng]{mildenhall2021nerf}
Ben Mildenhall, Pratul~P Srinivasan, Matthew Tancik, Jonathan~T Barron, Ravi Ramamoorthi, and Ren Ng.
\newblock Nerf: Representing scenes as neural radiance fields for view synthesis.
\newblock \emph{Communications of the ACM}, 65\penalty0 (1):\penalty0 99--106, 2021.

\bibitem[Nekrasov et~al.(2021)Nekrasov, Schult, Litany, Leibe, and Engelmann]{mix3d}
Alexey Nekrasov, Jonas Schult, Or Litany, Bastian Leibe, and Francis Engelmann.
\newblock Mix3d: Out-of-context data augmentation for 3d scenes.
\newblock In \emph{2021 international conference on 3d vision (3dv)}, pages 116--125. IEEE, 2021.

\bibitem[Pan et~al.(2023)Pan, Tang, Wang, and Shan]{tap}
Ting Pan, Lulu Tang, Xinlong Wang, and Shiguang Shan.
\newblock Tokenize anything via prompting.
\newblock \emph{arXiv preprint arXiv:2312.09128}, 2023.

\bibitem[Pan et~al.(2021)Pan, Xia, Song, Li, and Huang]{pan20213d}
Xuran Pan, Zhuofan Xia, Shiji Song, Li~Erran Li, and Gao Huang.
\newblock 3d object detection with pointformer.
\newblock In \emph{Proceedings of the IEEE/CVF conference on computer vision and pattern recognition}, pages 7463--7472, 2021.

\bibitem[Park et~al.(2020)Park, Kim, and Lee]{VR}
Kyeong-Beom Park, Minseok Kim, and Jae~Yeol Lee.
\newblock Deep learning-based smart task assistance in wearable augmented reality.
\newblock \emph{Robotics and Computer-Integrated Manufacturing}, page 101887, 2020.

\bibitem[Peng et~al.(2023{\natexlab{a}})Peng, Genova, Jiang, Tagliasacchi, Pollefeys, Funkhouser, et~al.]{openscene}
Songyou Peng, Kyle Genova, Chiyu Jiang, Andrea Tagliasacchi, Marc Pollefeys, Thomas Funkhouser, et~al.
\newblock Openscene: 3d scene understanding with open vocabularies.
\newblock In \emph{IEEE Conf. Comput. Vis. Pattern Recog.}, pages 815--824, 2023{\natexlab{a}}.

\bibitem[Peng et~al.(2023{\natexlab{b}})Peng, Genova, Jiang, Tagliasacchi, Pollefeys, Funkhouser, et~al.]{peng2023openscene}
Songyou Peng, Kyle Genova, Chiyu Jiang, Andrea Tagliasacchi, Marc Pollefeys, Thomas Funkhouser, et~al.
\newblock Openscene: 3d scene understanding with open vocabularies.
\newblock In \emph{Proceedings of the IEEE/CVF conference on computer vision and pattern recognition}, pages 815--824, 2023{\natexlab{b}}.

\bibitem[Qi et~al.(2017{\natexlab{a}})Qi, Su, Mo, and Guibas]{qi2017pointnet}
Charles~R Qi, Hao Su, Kaichun Mo, and Leonidas~J Guibas.
\newblock Pointnet: Deep learning on point sets for 3d classification and segmentation.
\newblock In \emph{Proceedings of the IEEE conference on computer vision and pattern recognition}, pages 652--660, 2017{\natexlab{a}}.

\bibitem[Qi et~al.(2017{\natexlab{b}})Qi, Yi, Su, and Guibas]{pointnet++}
Charles~Ruizhongtai Qi, Li Yi, Hao Su, and Leonidas~J Guibas.
\newblock Pointnet++: Deep hierarchical feature learning on point sets in a metric space.
\newblock \emph{NeurIPS}, 2017{\natexlab{b}}.

\bibitem[Qi et~al.(2017{\natexlab{c}})Qi, Yi, Su, and Guibas]{qi2017pointnet++}
Charles~Ruizhongtai Qi, Li Yi, Hao Su, and Leonidas~J Guibas.
\newblock Pointnet++: Deep hierarchical feature learning on point sets in a metric space.
\newblock \emph{Advances in neural information processing systems}, 30, 2017{\natexlab{c}}.

\bibitem[Qin et~al.(2024{\natexlab{a}})Qin, Li, Zhou, Wang, and Pfister]{langsplat}
Minghan Qin, Wanhua Li, Jiawei Zhou, Haoqian Wang, and Hanspeter Pfister.
\newblock Langsplat: 3d language gaussian splatting.
\newblock In \emph{Proceedings of the IEEE/CVF Conference on Computer Vision and Pattern Recognition}, pages 20051--20060, 2024{\natexlab{a}}.

\bibitem[Qin et~al.(2024{\natexlab{b}})Qin, Li, Zhou, Wang, and Pfister]{qin2024langsplat}
Minghan Qin, Wanhua Li, Jiawei Zhou, Haoqian Wang, and Hanspeter Pfister.
\newblock Langsplat: 3d language gaussian splatting.
\newblock In \emph{Proceedings of the IEEE/CVF Conference on Computer Vision and Pattern Recognition}, pages 20051--20060, 2024{\natexlab{b}}.

\bibitem[Radford et~al.(2021{\natexlab{a}})Radford, Kim, Hallacy, Ramesh, Goh, Agarwal, Sastry, Askell, Mishkin, Clark, et~al.]{clip}
Alec Radford, Jong~Wook Kim, Chris Hallacy, Aditya Ramesh, Gabriel Goh, Sandhini Agarwal, Girish Sastry, Amanda Askell, Pamela Mishkin, Jack Clark, et~al.
\newblock Learning transferable visual models from natural language supervision.
\newblock pages 8748--8763. PMLR, 2021{\natexlab{a}}.

\bibitem[Radford et~al.(2021{\natexlab{b}})Radford, Kim, Hallacy, Ramesh, Goh, Agarwal, Sastry, Askell, Mishkin, Clark, et~al.]{radford2021learning}
Alec Radford, Jong~Wook Kim, Chris Hallacy, Aditya Ramesh, Gabriel Goh, Sandhini Agarwal, Girish Sastry, Amanda Askell, Pamela Mishkin, Jack Clark, et~al.
\newblock Learning transferable visual models from natural language supervision.
\newblock In \emph{International conference on machine learning}, pages 8748--8763. PMLR, 2021{\natexlab{b}}.

\bibitem[Rombach et~al.(2021)Rombach, Blattmann, Lorenz, Esser, and Ommer]{stablediffusion}
Robin Rombach, Andreas Blattmann, Dominik Lorenz, Patrick Esser, and Björn Ommer.
\newblock High-resolution image synthesis with latent diffusion models, 2021.

\bibitem[Rombach et~al.(2022)Rombach, Blattmann, Lorenz, Esser, and Ommer]{rombach2022high}
Robin Rombach, Andreas Blattmann, Dominik Lorenz, Patrick Esser, and Bj{\"o}rn Ommer.
\newblock High-resolution image synthesis with latent diffusion models.
\newblock In \emph{Proceedings of the IEEE/CVF conference on computer vision and pattern recognition}, pages 10684--10695, 2022.

\bibitem[Schult et~al.(2022)Schult, Engelmann, Hermans, Litany, Tang, and Leibe]{schult2022mask3d}
Jonas Schult, Francis Engelmann, Alexander Hermans, Or Litany, Siyu Tang, and Bastian Leibe.
\newblock Mask3d for 3d semantic instance segmentation.
\newblock \emph{arXiv preprint arXiv:2210.03105}, 2022.

\bibitem[Seita et~al.(2023)Seita, Wang, Shetty, Erickson, and Held]{robot}
Daniel Seita, Yufei Wang, Sarthak~J Shetty, Zackory Erickson, and David Held.
\newblock Toolflownet: Robotic manipulation with tools via predicting tool flow from point clouds.
\newblock In \emph{Conference on Robot Learning}, pages 1038--1049. PMLR, 2023.

\bibitem[Shen et~al.(2023)Shen, Geng, Yuan, Lin, Liu, Wang, Hu, Zheng, and Guo]{shen2023v}
Yichao Shen, Zigang Geng, Yuhui Yuan, Yutong Lin, Ze Liu, Chunyu Wang, Han Hu, Nanning Zheng, and Baining Guo.
\newblock V-detr: Detr with vertex relative position encoding for 3d object detection.
\newblock \emph{arXiv preprint arXiv:2308.04409}, 2023.

\bibitem[Siddiqui et~al.(2023)Siddiqui, Porzi, Bul{\`o}, M{\"u}ller, Nie{\ss}ner, Dai, and Kontschieder]{panopticlifting}
Yawar Siddiqui, Lorenzo Porzi, Samuel~Rota Bul{\`o}, Norman M{\"u}ller, Matthias Nie{\ss}ner, Angela Dai, and Peter Kontschieder.
\newblock Panoptic lifting for 3d scene understanding with neural fields.
\newblock In \emph{IEEE Conf. Comput. Vis. Pattern Recog.}, pages 9043--9052, 2023.

\bibitem[Sun et~al.(2023)Sun, Qing, Tan, and Xu]{sun2023superpoint}
Jiahao Sun, Chunmei Qing, Junpeng Tan, and Xiangmin Xu.
\newblock Superpoint transformer for 3d scene instance segmentation.
\newblock In \emph{Proceedings of the AAAI Conference on Artificial Intelligence}, pages 2393--2401, 2023.

\bibitem[Tan and Le(2019)]{tan2019efficientnet}
Mingxing Tan and Quoc Le.
\newblock Efficientnet: Rethinking model scaling for convolutional neural networks.
\newblock In \emph{International conference on machine learning}, pages 6105--6114. PMLR, 2019.

\bibitem[Tang et~al.(2022)Tang, Chen, Han, Liao, Ru, and Wu]{tang2022bi}
Zaizuo Tang, Guangzhu Chen, Yinhe Han, Xiaojuan Liao, Qingjun Ru, and Yuanyuan Wu.
\newblock Bi-stage multi-modal 3d instance segmentation method for production workshop scene.
\newblock \emph{Engineering Applications of Artificial Intelligence}, 112:\penalty0 104858, 2022.

\bibitem[Tarvainen and Valpola(2017)]{tarvainen2017mean}
Antti Tarvainen and Harri Valpola.
\newblock Mean teachers are better role models: Weight-averaged consistency targets improve semi-supervised deep learning results.
\newblock \emph{Advances in neural information processing systems}, 30, 2017.

\bibitem[Thomas et~al.(2019)Thomas, Qi, Goulette, and Guibas]{kpconv}
Hugues Thomas, Charles~R Qi, Fran{\c{c}}ois Goulette, and Leonidas~J Guibas.
\newblock Kpconv: Flexible and deformable convolution for point clouds.
\newblock In \emph{ICCV}, pages 6411--6420, 2019.

\bibitem[Vu et~al.(2022)Vu, Kim, Luu, Nguyen, and Yoo]{vu2022softgroup}
Thang Vu, Kookhoi Kim, Tung~M Luu, Thanh Nguyen, and Chang~D Yoo.
\newblock Softgroup for 3d instance segmentation on point clouds.
\newblock In \emph{Proceedings of the IEEE/CVF Conference on Computer Vision and Pattern Recognition}, pages 2708--2717, 2022.

\bibitem[Wang et~al.(2022)Wang, Chen, and Yang]{dmnerf}
Bing Wang, Lu Chen, and Bo Yang.
\newblock Dm-nerf: 3d scene geometry decomposition and manipulation from 2d images.
\newblock \emph{arXiv preprint arXiv:2208.07227}, 2022.

\bibitem[Wang et~al.(2023)Wang, Zhang, Song, Bi, Zhang, Wei, Tang, Yang, Li, Jia, et~al.]{wang2023multi}
Li Wang, Xinyu Zhang, Ziying Song, Jiangfeng Bi, Guoxin Zhang, Haiyue Wei, Liyao Tang, Lei Yang, Jun Li, Caiyan Jia, et~al.
\newblock Multi-modal 3d object detection in autonomous driving: A survey and taxonomy.
\newblock \emph{IEEE Transactions on Intelligent Vehicles}, 8\penalty0 (7):\penalty0 3781--3798, 2023.

\bibitem[Wang et~al.(2024{\natexlab{a}})Wang, Wang, Li, Zhang, Lei, and Zhang]{GGSD}
Pengfei Wang, Yuxi Wang, Shuai Li, Zhaoxiang Zhang, Zhen Lei, and Lei Zhang.
\newblock Open vocabulary 3d scene understanding via geometry guided self-distillation.
\newblock In \emph{European Conference on Computer Vision}, pages 442--460. Springer, 2024{\natexlab{a}}.

\bibitem[Wang et~al.(2024{\natexlab{b}})Wang, Wang, Li, Zhang, Lei, and Zhang]{wang2024open}
Pengfei Wang, Yuxi Wang, Shuai Li, Zhaoxiang Zhang, Zhen Lei, and Lei Zhang.
\newblock Open vocabulary 3d scene understanding via geometry guided self-distillation.
\newblock In \emph{European Conference on Computer Vision}, pages 442--460. Springer, 2024{\natexlab{b}}.

\bibitem[Wu et~al.(2019)Wu, Qi, and Fuxin]{pointconv}
Wenxuan Wu, Zhongang Qi, and Li Fuxin.
\newblock Pointconv: Deep convolutional networks on 3d point clouds.
\newblock In \emph{CVPR}, pages 9621--9630, 2019.

\bibitem[Wu et~al.(2023)Wu, Zhao, Shou, Zhou, and Shen]{diffumask}
Weijia Wu, Yuzhong Zhao, Mike~Zheng Shou, Hong Zhou, and Chunhua Shen.
\newblock Diffumask: Synthesizing images with pixel-level annotations for semantic segmentation using diffusion models.
\newblock In \emph{Proceedings of the IEEE/CVF International Conference on Computer Vision}, pages 1206--1217, 2023.

\bibitem[Wu et~al.(2024{\natexlab{a}})Wu, Jiang, Ouyang, He, and Zhao]{PTv3}
Xiaoyang Wu, Li Jiang, Wanli Ouyang, Tong He, and Hengshuang Zhao.
\newblock Point transformer v3: Simpler faster stronger.
\newblock In \emph{CVPR}, 2024{\natexlab{a}}.

\bibitem[Wu et~al.(2024{\natexlab{b}})Wu, Jiang, Wang, Liu, Liu, Qiao, Ouyang, He, and Zhao]{wu2024point}
Xiaoyang Wu, Li Jiang, Peng-Shuai Wang, Zhijian Liu, Xihui Liu, Yu Qiao, Wanli Ouyang, Tong He, and Hengshuang Zhao.
\newblock Point transformer v3: Simpler faster stronger.
\newblock In \emph{Proceedings of the IEEE/CVF Conference on Computer Vision and Pattern Recognition}, pages 4840--4851, 2024{\natexlab{b}}.

\bibitem[Xu et~al.(2023{\natexlab{a}})Xu, Liu, Vahdat, Byeon, Wang, and De~Mello]{odise}
Jiarui Xu, Sifei Liu, Arash Vahdat, Wonmin Byeon, Xiaolong Wang, and Shalini De~Mello.
\newblock {Open-Vocabulary Panoptic Segmentation with Text-to-Image Diffusion Models}.
\newblock \emph{arXiv preprint arXiv:2303.04803}, 2023{\natexlab{a}}.

\bibitem[Xu et~al.(2023{\natexlab{b}})Xu, Zhang, Wei, Hu, and Bai]{xu2023side}
Mengde Xu, Zheng Zhang, Fangyun Wei, Han Hu, and Xiang Bai.
\newblock Side adapter network for open-vocabulary semantic segmentation.
\newblock In \emph{Proceedings of the IEEE/CVF Conference on Computer Vision and Pattern Recognition}, pages 2945--2954, 2023{\natexlab{b}}.

\bibitem[Yang et~al.(2024)Yang, Ding, Deng, Wang, and Qi]{regionplc}
Jihan Yang, Runyu Ding, Weipeng Deng, Zhe Wang, and Xiaojuan Qi.
\newblock Regionplc: Regional point-language contrastive learning for open-world 3d scene understanding.
\newblock In \emph{Proceedings of the IEEE/CVF Conference on Computer Vision and Pattern Recognition}, 2024.

\bibitem[Ye et~al.(2023)Ye, Zhou, Chen, Xie, Wang, Wang, and Foroosh]{lidarmultinet}
Dongqiangzi Ye, Zixiang Zhou, Weijia Chen, Yufei Xie, Yu Wang, Panqu Wang, and Hassan Foroosh.
\newblock Lidarmultinet: Towards a unified multi-task network for lidar perception.
\newblock In \emph{Proceedings of the AAAI Conference on Artificial Intelligence}, pages 3231--3240, 2023.

\bibitem[Ye et~al.(2024)Ye, Danelljan, Yu, and Ke]{ye2024gaussian}
Mingqiao Ye, Martin Danelljan, Fisher Yu, and Lei Ke.
\newblock Gaussian grouping: Segment and edit anything in 3d scenes.
\newblock In \emph{European Conference on Computer Vision}, pages 162--179. Springer, 2024.

\bibitem[Zeng et~al.(2021)Zeng, Qian, Zhu, Hou, Yuan, and He]{zeng2021corrnet3d}
Yiming Zeng, Yue Qian, Zhiyu Zhu, Junhui Hou, Hui Yuan, and Ying He.
\newblock Corrnet3d: Unsupervised end-to-end learning of dense correspondence for 3d point clouds.
\newblock In \emph{Proceedings of the IEEE/CVF Conference on Computer Vision and Pattern Recognition}, pages 6052--6061, 2021.

\bibitem[Zhang et~al.(2023)Zhang, Dong, and Ma]{clipfo3d}
Junbo Zhang, Runpei Dong, and Kaisheng Ma.
\newblock Clip-fo3d: Learning free open-world 3d scene representations from 2d dense clip.
\newblock In \emph{Proceedings of the IEEE/CVF International Conference on Computer Vision}, pages 2048--2059, 2023.

\bibitem[Zhang et~al.(2022)Zhang, Guo, Zhang, Li, Miao, Cui, Qiao, Gao, and Li]{zhang2022pointclip}
Renrui Zhang, Ziyu Guo, Wei Zhang, Kunchang Li, Xupeng Miao, Bin Cui, Yu Qiao, Peng Gao, and Hongsheng Li.
\newblock Pointclip: Point cloud understanding by clip.
\newblock In \emph{Proceedings of the IEEE/CVF conference on computer vision and pattern recognition}, pages 8552--8562, 2022.

\bibitem[Zhao et~al.(2021)Zhao, Jiang, Jia, Torr, and Koltun]{zhao2021point}
Hengshuang Zhao, Li Jiang, Jiaya Jia, Philip~HS Torr, and Vladlen Koltun.
\newblock Point transformer.
\newblock In \emph{Proceedings of the IEEE/CVF international conference on computer vision}, pages 16259--16268, 2021.

\bibitem[Zhi et~al.(2021)Zhi, Laidlow, Leutenegger, and Davison]{semanticnerf}
Shuaifeng Zhi, Tristan Laidlow, Stefan Leutenegger, and Andrew~J Davison.
\newblock In-place scene labelling and understanding with implicit scene representation.
\newblock In \emph{IEEE Conf. Comput. Vis. Pattern Recog.}, pages 15838--15847, 2021.

\bibitem[Zhou et~al.(2024)Zhou, Chang, Jiang, Fan, Zhu, Xu, Chari, You, Wang, and Kadambi]{feature3dgs}
Shijie Zhou, Haoran Chang, Sicheng Jiang, Zhiwen Fan, Zehao Zhu, Dejia Xu, Pradyumna Chari, Suya You, Zhangyang Wang, and Achuta Kadambi.
\newblock Feature 3dgs: Supercharging 3d gaussian splatting to enable distilled feature fields.
\newblock In \emph{Proceedings of the IEEE/CVF Conference on Computer Vision and Pattern Recognition}, pages 21676--21685, 2024.

\bibitem[Zhou and Tuzel(2018)]{zhou2018voxelnet}
Yin Zhou and Oncel Tuzel.
\newblock Voxelnet: End-to-end learning for point cloud based 3d object detection.
\newblock In \emph{Proceedings of the IEEE conference on computer vision and pattern recognition}, pages 4490--4499, 2018.

\bibitem[Zhu et~al.(2024{\natexlab{a}})Zhu, Zhou, Xing, Zhao, Xu, Liang, Hauptmann, Liu, and Gallagher]{diff2scene}
Xiaoyu Zhu, Hao Zhou, Pengfei Xing, Long Zhao, Hao Xu, Junwei Liang, Alexander Hauptmann, Ting Liu, and Andrew Gallagher.
\newblock Open-vocabulary 3d semantic segmentation with text-to-image diffusion models.
\newblock In \emph{European Conference on Computer Vision}, pages 357--375. Springer, 2024{\natexlab{a}}.

\bibitem[Zhu et~al.(2024{\natexlab{b}})Zhu, Zhou, Xing, Zhao, Xu, Liang, Hauptmann, Liu, and Gallagher]{zhu2024open}
Xiaoyu Zhu, Hao Zhou, Pengfei Xing, Long Zhao, Hao Xu, Junwei Liang, Alexander Hauptmann, Ting Liu, and Andrew Gallagher.
\newblock Open-vocabulary 3d semantic segmentation with text-to-image diffusion models.
\newblock In \emph{European Conference on Computer Vision}, pages 357--375. Springer, 2024{\natexlab{b}}.

\bibitem[Zou et~al.(2023)Zou, Yang, Zhang, Li, Li, Wang, Wang, Gao, and Lee]{SEEM}
Xueyan Zou, Jianwei Yang, Hao Zhang, Feng Li, Linjie Li, Jianfeng Wang, Lijuan Wang, Jianfeng Gao, and Yong~Jae Lee.
\newblock Segment everything everywhere all at once.
\newblock \emph{Advances in neural information processing systems}, 36:\penalty0 19769--19782, 2023.

\bibitem[Zou et~al.(2024)Zou, Yang, Zhang, Li, Li, Wang, Wang, Gao, and Lee]{zou2024segment}
Xueyan Zou, Jianwei Yang, Hao Zhang, Feng Li, Linjie Li, Jianfeng Wang, Lijuan Wang, Jianfeng Gao, and Yong~Jae Lee.
\newblock Segment everything everywhere all at once.
\newblock \emph{Advances in Neural Information Processing Systems}, 36, 2024.

\end{thebibliography}
}
\newpage
\section{Appendix}
In this \textbf{supplementary material}, we first provide implementation details in Sec. \ref{sec:A}.
Then, we supply additional qualitative results in Sec. \ref{sec:B}.

\subsection{Implementation Details}
\label{sec:A}

\paragraph{Training setting}

we provide full information of the training configuration as shown in Tab. \ref{tab:1}. Specifically, we adopt Adam \cite{adam} as the optimizer with a base learning rate of $1e-4$. The learning scheduler adjusts the learning rate linearly to $1e-5$ throughout the whole process. The weight decay is set to 0. We use a batch size of 12 and 8 for indoor scenes and outdoor scenes respectively to train for 100 epoches in total. Besides, the voxel size is set to 2cm and 5cm respectively for indoor scenes and outdoor scenes.

\begin{table}[h]
    \centering
    \setlength{\tabcolsep}{0.5mm}
    \begin{tabular}{lclc}
    \toprule
    \multicolumn{2}{c}{ScanNet v2 \cite{scannet} / Matterport3D \cite{matterport3d}} &\multicolumn{2}{c}{nuScenes \cite{nuscenes}}  \\
    \cmidrule(lr){1-2} \cmidrule(lr){3-4} 
    Config          &Value      &Config         &Value            \\\midrule
    optimizer       &Adam \cite{adam}      &optimizer      &Adam \cite{adam}              \\
    scheduler       &Linear     &scheduler      &Linear           \\
    base lr         &1e-4       &base lr        &1e-4            \\
    weight decay    &0       &weight decay   &0            \\
    batch size      &12         &batch size     &8               \\
    epochs          &100        &epochs         &100               \\
    voxel size      &2cm     &voxel size     &5cm     \\
    \bottomrule
    \end{tabular}
    \caption{\textbf{Training settings.} Here we list the training settings for both indoor scenes and outdoor scenes.}
    \label{tab:1}
\end{table}

\paragraph{Model architecture}
We adopt MinkowskiNet18A \cite{minknet} to be the architecture of the 3D distilled model, which is consistent with OpenScene \cite{openscene}. Besides, the input to the 3D distilled model is the pure point cloud without color or other attributes. 

\begin{table}[!t]
    \centering
    \footnotesize
    \setlength{\tabcolsep}{0.25cm}
    \resizebox{0.48\textwidth}{!}{

    \begin{tabular}{l|l}
        \toprule
        nuScenes 16 labels & OpenScene's pre-defined labels\\
        \midrule
        barrier & barrier, barricade\\
        bicycle & bicycle\\
        bus & bus\\
        car  & car\\
        construction vehicle  &  \parbox[t]{4cm}{bulldozer, excavator, concrete mixer, crane, dump truck}\\
        motorcycle  & motorcycle\\
        pedestrian  & pedestrian, person\\
        traffic cone  & traffic cone\\
        trailer  & \parbox[t]{4cm}{trailer, semi trailer, cargo container, shipping container, freight container}\\
        truck  & truck\\
        driveable surface  & road\\
        other flat  & \parbox[t]{4cm}{curb, traffic island, traffic median}\\
        sidewalk  & sidewalk\\
        terrain  & \parbox[t]{4cm}{grass, grassland, lawn, meadow, turf, sod}\\
        manmade  & building, wall, pole, awning\\
        vegetation & \parbox[t]{4cm}{tree, trunk, tree trunk, bush, shrub, plant, flower, woods}\\
        \bottomrule
    \end{tabular}}
    \caption{
        \textbf{Label Mappings for nuScenes 16 Classes.} Here we list the total 43 pre-defined non-ambiguous class names corresponding to the 16 nuScenes classes. 
        }
\label{tab:2}
\end{table}

\paragraph{nuScenes inference}
As some category names in nuScenes \cite{nuscenes} have ambiguous meanings, e.g., ``drivable surface" and ``other flat", we follow OpenScene \cite{openscene} to pre-define some detailed category names that have clear meanings, and then map the predictions from these pre-defined categories back to the original categories. The original categories and the pre-defined categories are shown in Tab. \ref{tab:2}.

\begin{figure*}[t]
  \centering
  \includegraphics[scale=0.3]{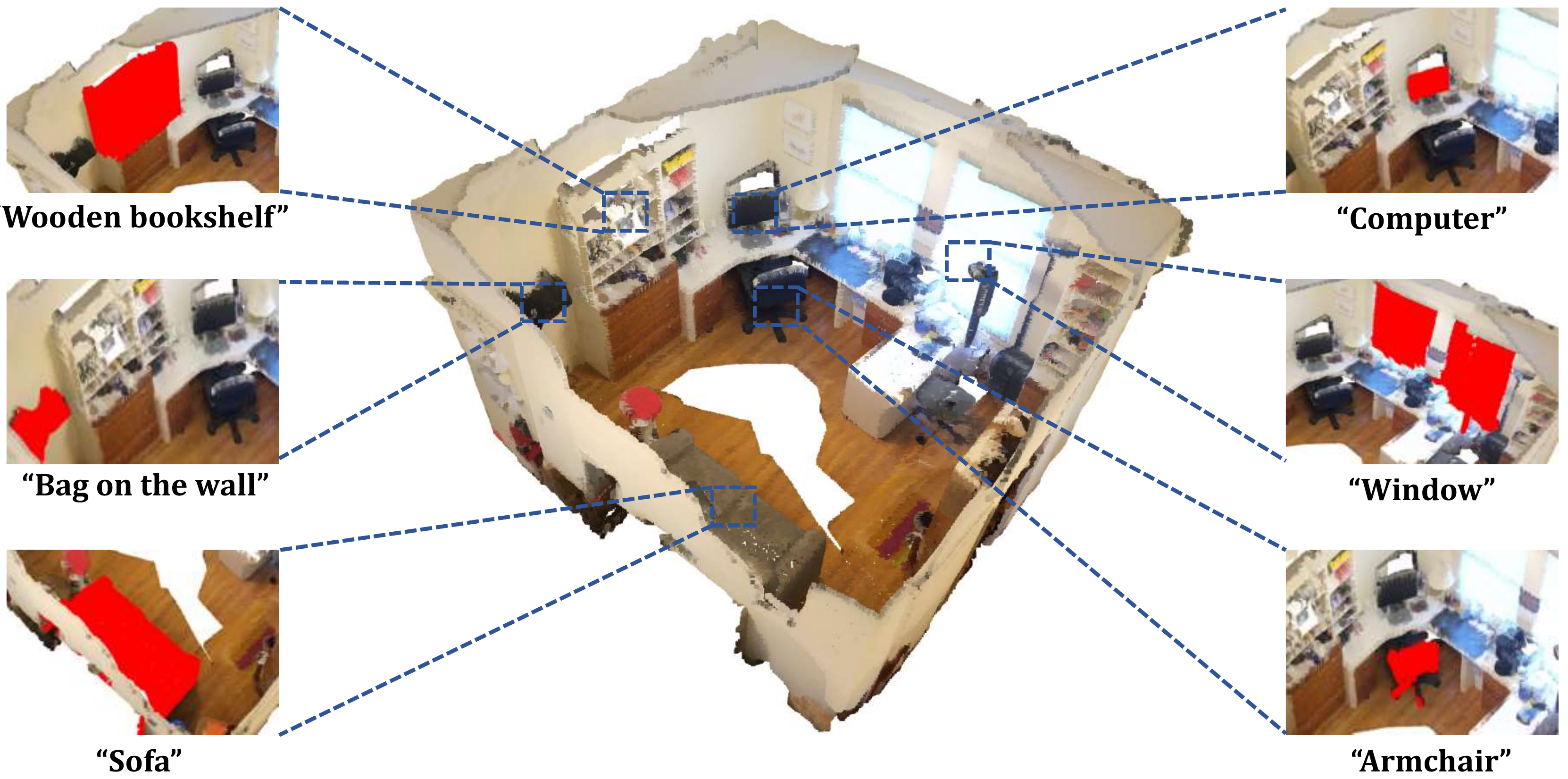}
  \caption{\textbf{Querying about different objects in a scene.} The scene is collected from ScanNet v2. Red indicate the queried parts that match the text description.}
  \label{fig:1}
\end{figure*}

\paragraph{Multi-view feature fusion}
Multi-view feature fusion is to aggregate the 2D image features onto 3D points through pixel-point correspondence. Our multi-view feature fusion strategy is exactly the same with OpenScene's \cite{openscene}. Specifically, for Matterport3D \cite{matterport3d} and nuScenes \cite{nuscenes}, we aggregate features of all images from every scene onto the 3D point, while we only sample 1 image out of every 20 video frames and fuse them for ScanNet \cite{scannet}. 
Besides, we conduct occlusion tests for ScanNet v2 \cite{scannet} and Matterport3D \cite{matterport3d} as they provide depth information of each image, which guarantees that a pixel is only connected to a visible surface point. Specifically, for a single point, we calculate its distance between it and its corresponding pixel. If the difference between the distance and the pixel's depth value $D$ is smaller than a threshold $\sigma$, we can connect the point to this pixel. Otherwise, we do not project the pixel's features onto the point cloud. We set $\sigma=0.2D$ for ScanNet v2 \cite{scannet} and $\sigma=0.02D$ for Matterport3D \cite{matterport3d}, which is consistent with OpenScene \cite{openscene}.  

\paragraph{Superpoint generation}
We compute superpoints only for indoor datasets ScanNet v2 \cite{scannet} and Matterport3D \cite{matterport3d}. Specifically, we use the mesh data provided by ScanNet v2 \cite{scannet} and Matterport3D \cite{matterport3d} as input. We extract superpoints from the mesh by performing a graph-based algorithm \cite{effiseg} on the computed mesh normals. For nuScenes \cite{nuscenes}, we do not compute any superpoint and treat every single point as a superpoint since ourdoor point clouds are normally dominated by ``road", making it hard to extract superpoints.

\paragraph{Prompt engineering}
When extracting text features during inference, we apply a simple prompt engineering that modifies the class name ``XX" to ``a XX in a scene" to generate a better performance, which is proven by OpenScene \cite{openscene}. 
Besides, when synthesizing images in Sec 3.2 in main paper, we apply another prompt engineering that modifies the class name ``XX" to ``a good photo of XX" to obtain high quality images.

\paragraph{Pre-built vocabulary}
We construct two pre-built vocabulary (Sec 3.2) for indoor scenes and outdoor scenes respectively, as shown in Tab. \ref{tab:3}.
\begin{table}[h]
    \centering
    \footnotesize
    \setlength{\tabcolsep}{0.25cm}
    \resizebox{0.48\textwidth}{!}{

    \begin{tabular}{l|l}
        \toprule
        Indoor scenes & Outdoor scenes\\
        \midrule
        bookshelf, table, wall, & person, bus, wall, \\
        bathtub, sofa, ceiling,& grass, car, bicycle,\\
        door, bed, toilet, & crane, sky, tree,\\
        picture, desk, floor, & excavator, barricade, trailer,\\
        counter, shower curtain, sink& pavement, building, road,\\
        curtain, window, chair,& motorcycle, plant, truck,\\
        cabinet, refrigerator,&awning, container, lawn,\\
        & traffi cone, bulldozer\\

        \bottomrule
    \end{tabular}}
    \caption{
        \textbf{Pre-built vocabulary.} Here we give the detail of constructed pre-built vocabulary for indoor scenes and outdoor scenes respectiely.
        }
\label{tab:3}
\end{table}

\subsection{Additional Qualitative Results}
\label{sec:B}

\begin{figure*}[t]
  \centering
  \includegraphics[scale=0.25]{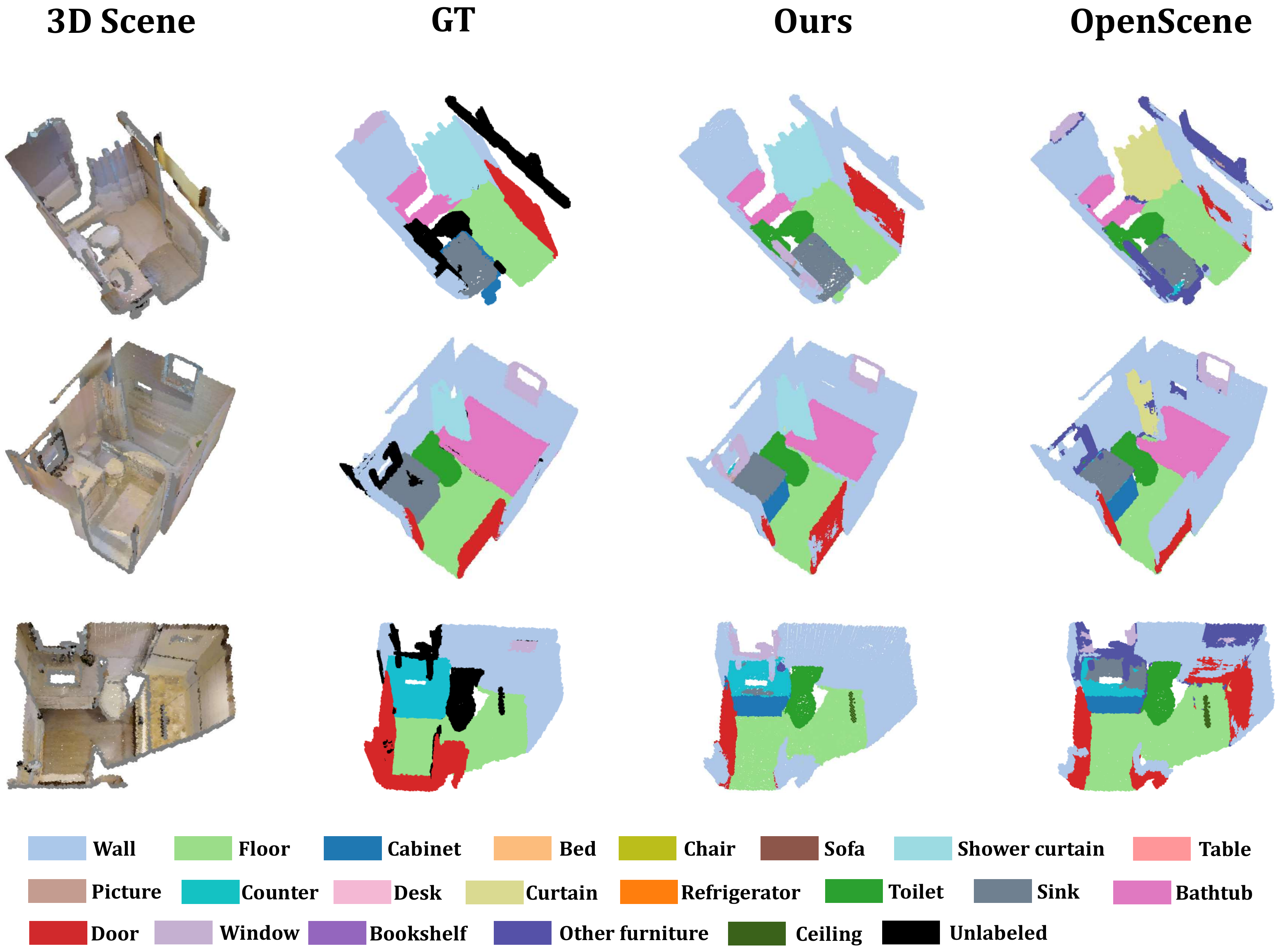}
  \caption{\textbf{Visualization results.} Semantic segmentation results of \name{} on Matterport3D \cite{matterport3d}.}
  \label{fig:2}
  \vspace{-10pt}
\end{figure*}
\paragraph{Querying objects in a scene}
We display a visualization of querying about different objects in a scene as shown in Fig. \ref{fig:1}. First, we adopt the 3D distilled model to output per-point features. Then we use different query texts and encode them with CLIP to obtain text features. By computing the similarity between point features and text features, we denote points with high similarity as red.

\paragraph{Visualization on Matterport3D}

Visual Comparisons with OpenScene \cite{openscene} on semantic segmentation in Matterport3D \cite{matterport3d} are shown in Fig. \ref{fig:2}, which our proposed \name() effectively corrects some wrong predictions made by OpenScene \cite{openscene}. For, example, OpenScene \cite{openscene} misidentifies a shower curtain as a curtain, while \name{} can easily fix it.

\paragraph{Visualization on nuScenes}
\begin{figure*}[t]
  \centering
  \includegraphics[scale=0.25]{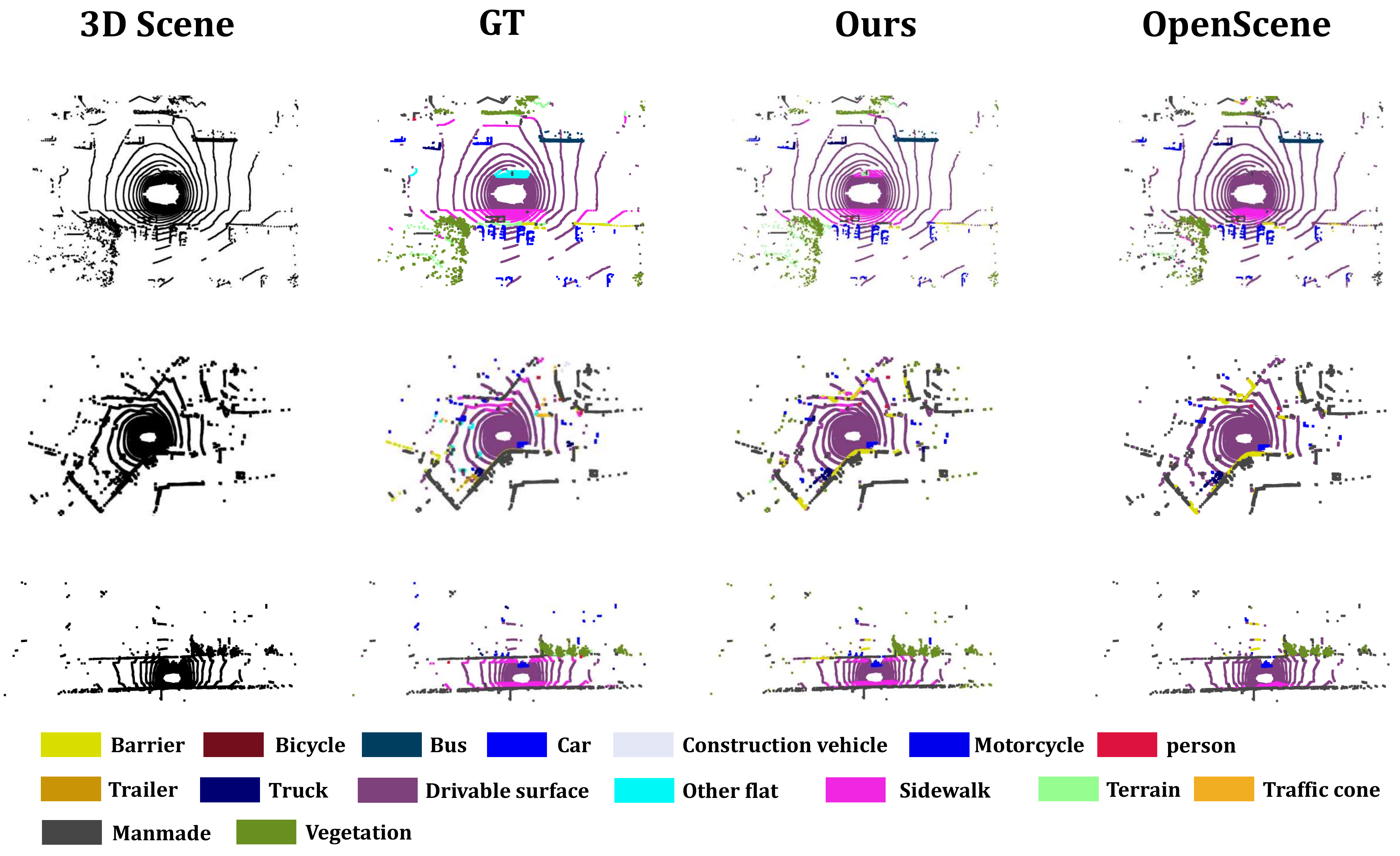}
  \caption{\textbf{Visualization results.} Semantic segmentation results of \name{} on nuScenes \cite{nuscenes}.}
  \label{fig:3}
\end{figure*}
Visual Comparisons with OpenScene \cite{openscene} on semantic segmentation in nuScenes \cite{nuscenes} is also shown are Fig. \ref{fig:3}.

\paragraph{Visualization on gaussian segmentation results}
\begin{figure*}[t]
  \centering
  \includegraphics[scale=0.25]{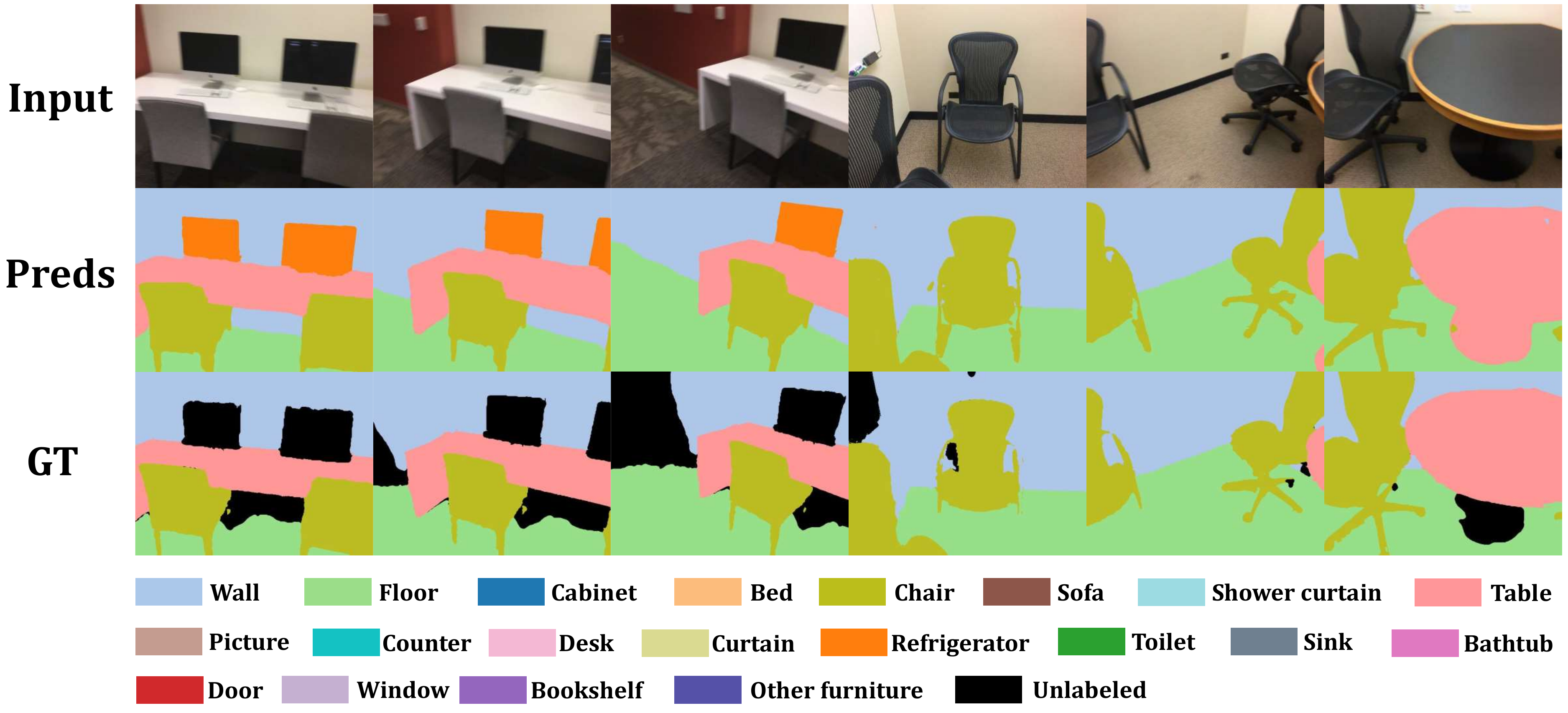}
  \caption{\textbf{Visualization results.} Gaussian semantic segmentation results of \name{} on ScanNet v2 \cite{scannet}.}
  \label{fig:4}
\end{figure*}
We also display the visualization of the gaussian segmentation on ScanNet v2 \cite{scannet} in Fig. \ref{fig:4}.

\end{document}